\documentclass[sigconf]{acmart}
\usepackage{tabularx}
\usepackage{array}
\usepackage{colortbl}
\usepackage{pifont}

\definecolor{DriveTeal}{HTML}{5AA7A7}
\definecolor{DriveMint}{HTML}{96D7C6}
\definecolor{DriveGreen}{HTML}{8AC94A}
\definecolor{DriveGold}{HTML}{E2D368}
\definecolor{DriveBlue}{HTML}{6C8CBF}

\newcommand{\cmark}{\textcolor{DriveGreen!65!black}{\ding{51}}}
\newcommand{\pmark}{\textcolor{DriveGold!75!black}{\ensuremath{\sim}}}
\newcommand{\xmark}{\textcolor{black!42}{\ding{55}}}

\usepackage{booktabs}
\usepackage{tikz}
\usetikzlibrary{positioning,arrows.meta}

\AtBeginDocument{%
  }

\setcopyright{acmlicensed}
\copyrightyear{2027}
\acmYear{2027}
\acmConference[KDD '27]{Proceedings of the 33rd ACM SIGKDD Conference on Knowledge Discovery and Data Mining}{August 2027}{TBD}
\acmBooktitle{Proceedings of the 33rd ACM SIGKDD Conference on Knowledge Discovery and Data Mining (KDD '27)}

\begin{document}

\title{DriveDNA: A Large-Scale Multimodal Naturalistic Driving Dataset and Benchmark for Driving Style Identification}

\author{Yuhang Wang}
\email{yuhangw@usf.edu}
\affiliation{%
  \institution{University of South Florida}
  \city{Tampa}
  \state{Florida}
  \country{USA}
}

\author{Lingyao Li}
\email{lingyaoli@arizona.edu}
\affiliation{%
  \institution{University of Arizona}
  \city{Tucson}
  \state{Arizona}
  \country{USA}
}

\author{Hao Zhou}
\email{haozhou1@usf.edu}
\affiliation{%
  \institution{University of South Florida}
  \city{Tampa}
  \state{Florida}
  \country{USA}
}

\begin{teaserfigure}
  \centering
  \includegraphics[width=0.85\textwidth]{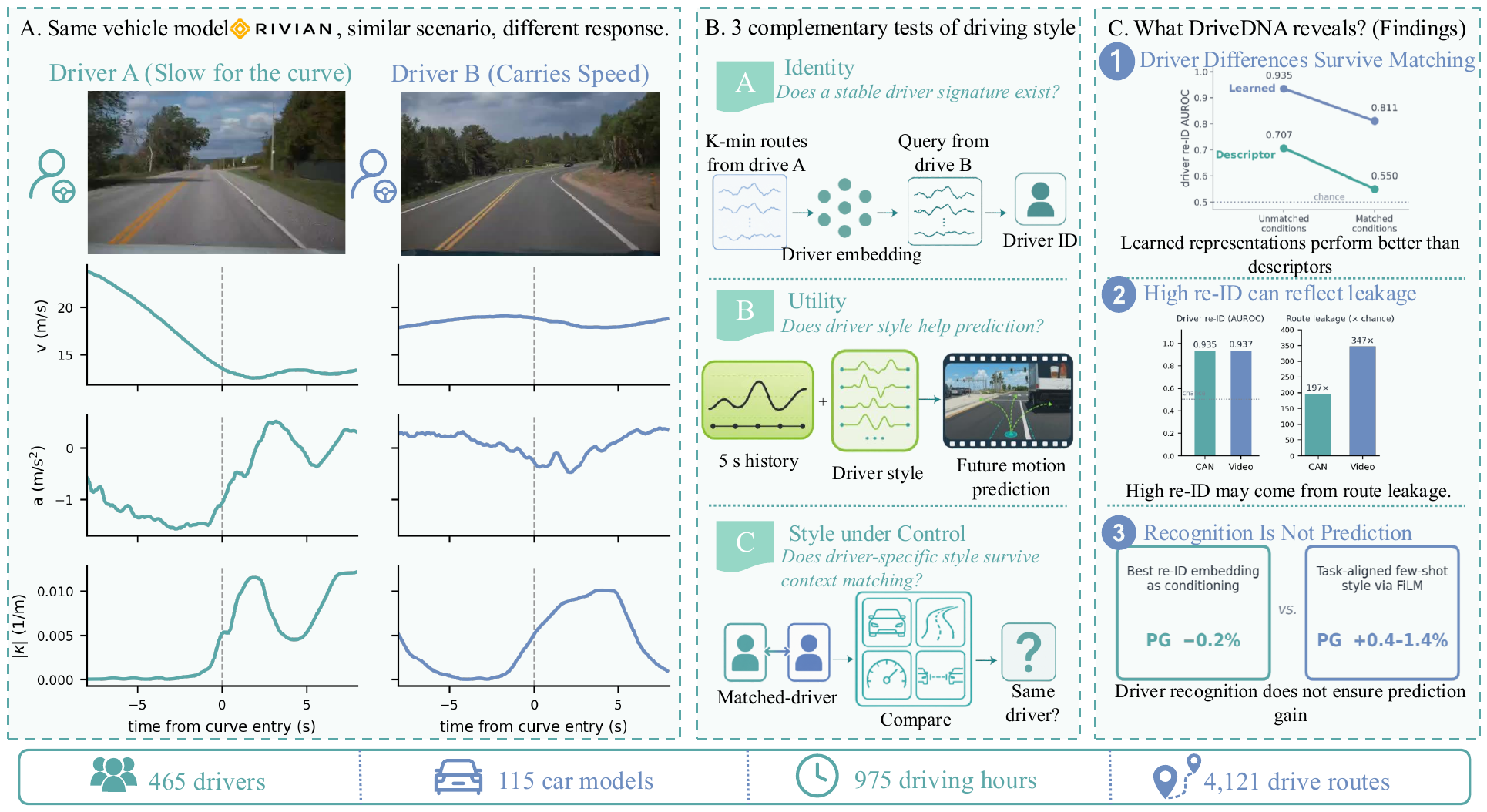}
  \caption{
    DriveDNA tests the identifiability of driver-specific behavioral patterns across vehicle models and traffic conditions. It benchmarks driving-style representations through few-shot driver identification, personalized behavior prediction, and condition-matched comparison.
  }
  \label{fig:teaser}
\end{teaserfigure}

\begin{abstract}
Driving style captures stable, driver-specific patterns in how a vehicle is driven. In naturalistic data, however, isolating this signal is difficult because drivers are observed in different vehicles, on different roads, and under different driving conditions. Models may therefore mistake vehicle- or situation-specific regularities for driver-specific style.
We therefore introduce DriveDNA, a large-scale naturalistic dataset and benchmark for personalized driving-style modeling, comprising 4,121 drives from 465 drivers across 115 vehicle models and totaling 975 hours of human-controlled driving at 10\,Hz with forward video.
The dataset is constructed from naturalistic driving logs from community drivers in their everyday use.
DriveDNA defines driving style as a consistent, driver-specific behavioral pattern in how a vehicle moves under similar driving conditions.
The benchmark evaluates this signal through three core tasks: few-shot driver re-identification, personalized behavior prediction, and condition-matched comparison.
It also provides behavioral annotations and 276{,}248 rule-generated maneuver events across six classes with large-scale human auditing.
We evaluate a broad suite of baselines spanning classical descriptors, supervised and self-supervised time-series encoders, multimodal fusion, probabilistic prediction, and zero-shot foundation models under a fixed multi-seed protocol.
Results show that learned representations substantially outperform classical descriptors on unseen drivers (AUROC .935 versus .707) and retain driver-specific information under matched driving conditions, while descriptor performance approaches chance.
Video-only models achieve comparable re-identification accuracy but exhibit severe route leakage, revealing that strong recognition performance may arise from contextual shortcuts rather than driving behavior.
Together, these findings show that reliable driving-style evaluation needs to assess both the behavioral value of learned representations and their robustness to vehicle, drive, and condition confounds.
\end{abstract}




\maketitle

\section{Introduction}
\label{sec:intro}

Driving behavior exhibits systematic individual variations: under comparable conditions, drivers differ in headway, speed choice, acceleration, lane-changing, and other control patterns~\cite{sagberg2015review,itkonen2020characterisation}.
These variations can serve as behavioral signatures supporting driver identification and motivate personalized driver assistance and autonomous planning~\cite{yang2020driver2vec,liao2025personalization}.
Recent benchmarks have begun to study personalization in controlled settings, simulation, and end-to-end planning~\cite{pdb,styledrive,pdbeval}; however, a benchmark that observes many drivers across many vehicles while explicitly controlling vehicle, route, and traffic-context confounds still remains missing~\cite{person2drive}.

Studying driving style in the wild requires naturalistic data at scale, as only everyday driving captures the diversity of drivers, vehicles, roads, and traffic conditions that personalization must ultimately handle~\cite{liao2025personalization}. 
Existing personalized-driving benchmarks~\cite{pdb,styledrive,person2drive,pdbeval} obtain cleaner comparisons through controlled collection or simulation, but at limited scale and heterogeneity. 
This scale and diversity, however, introduce a key methodological challenge: in naturalistic data, driver identity is often correlated with vehicle, route, and traffic exposure. 
A model may therefore appear to recognize a driver while actually recognizing the car they drive, the roads they frequent, or the situations they typically encounter. Rather than treating these confounds as a limitation to be acknowledged after evaluation, we make them a central target of the benchmark.

Addressing this challenge first requires an operational definition of ``driving style.'' We distinguish three layers: (i) driver input, such as pedal and steering commands, which is inconsistently observable and not directly comparable across vehicles; (ii) realized vehicle motion, including speed, acceleration, curvature, and headway, which provides the best-covered and most comparable behavioral space; and (iii) latent style, which is not directly observed but inferred as the stable driver-specific signal that remains after accounting for vehicle and context. We therefore operationalize driving style as stable, driver-specific patterns in realized vehicle motion under comparable contexts during human-controlled driving~\cite{sagberg2015review,itkonen2020characterisation}. This definition directly shapes both the construction of the dataset and the design of every benchmark task.

This paper introduces \textbf{DriveDNA}, a large-scale naturalistic dataset and benchmark for personalized driving-style identification. Its contributions are: 
\textbf{(1) A large, open, and community-driven naturalistic driving dataset.} DriveDNA contains 465 drivers, 115 vehicle models, 4{,}121 drives, and 975 hours of human-controlled driving at 10\,Hz with forward video and synchronized car control signals. Among them, 420 drivers share a vehicle model with at least one other driver, anchoring the within-nameplate comparisons, while 20 drivers appear on two or more models and anchor the cross-vehicle split. It is constructed from world-wide community drivers by removing automation-engaged segments and provides a driver/vehicle/drive hierarchy, per-drive vehicle parameters, and a human-audited maneuver-event layer;
\textbf{(2) The first benchmark designed to distinguish driver behavior from vehicle, drive, and driving-condition shortcuts.} We define three core tasks---driver re-identification, personalized behavior prediction, and condition-matched comparison---together with behavioral annotations, style-sensitive metrics, fixed evaluation procedures, and released data splits and evaluation harnesses;
\textbf{(3) Comprehensive baselines and diagnostics.} We evaluate thirty supervised, self-supervised, multimodal, and probabilistic baselines, accompanied by leakage probes that directly measure reliance on vehicle, route, and traffic-condition shortcuts. The dataset, splits, code, and evaluation are available at \href{https://huggingface.co/datasets/HenryYHW/DriveDNA}{Hugging Face}.

\section{Related Work}
\label{sec:related}

\textbf{Real-world driving data.}
Large public autonomous-driving datasets primarily support scene understanding and trajectory research, such as BDD100K~\cite{bdd100k} and nuScenes~\cite{nuscenes}. In particular, the Waymo Open Dataset~\cite{sun2020scalability} annotates road scenes and traffic participants, while highD~\cite{highd} provides naturalistic vehicle trajectories from an overhead view.
Resources centered more directly on human behavior include the controlled-access SHRP2 Naturalistic Driving Study~\cite{shrp2nds}, the 104-hour HDD corpus collected with one instrumented vehicle~\cite{hdd}, and comma2k19~\cite{comma2k19}, which releases dense camera, inertial, GNSS, and CAN logs from repeated highway commutes.
Focused naturalistic studies further characterize particular behaviors, such as car-following around automated vehicles~\cite{wen2022carfollowing} and longitudinal driving-style recognition across 44 drivers~\cite{lyu2022naturalistic}.
These resources establish the value of real-world observations, but are not organized as a public benchmark combining persistent driver identities, heterogeneous vehicles, human-only curation, and explicit vehicle, drive, and traffic-condition control.
Our proposed dataset \textbf{DriveDNA} targets this intersection, as listed in Table~\ref{tab:related_comparison}.

\begin{table*}[t]
\centering
\caption{
Representative datasets and benchmarks for driver behavior and personalized driving.
``ID'' denotes persistent driver identity across drives; \pmark{} indicates partial, indirect, or inherited support.
}
\label{tab:related_comparison}
\setlength{\tabcolsep}{2.0pt}
\renewcommand{\arraystretch}{1.04}
\scriptsize
\begin{tabularx}{\textwidth}{
  >{\raggedright\arraybackslash}p{0.142\textwidth}
  >{\centering\arraybackslash}p{0.060\textwidth}
  >{\raggedright\arraybackslash}p{0.128\textwidth}
  >{\raggedright\arraybackslash}p{0.120\textwidth}
  >{\centering\arraybackslash}p{0.043\textwidth}
  >{\centering\arraybackslash}p{0.047\textwidth}
  >{\centering\arraybackslash}p{0.050\textwidth}
  >{\raggedright\arraybackslash}X
  >{\raggedright\arraybackslash}p{0.147\textwidth}}
\toprule
\rowcolor{DriveTeal}
\color{white}\textbf{Work} &
\color{white}\textbf{Setting} &
\color{white}\textbf{Scale} &
\color{white}\textbf{Modalities} &
\color{white}\shortstack{\textbf{Persistent}\\\textbf{ID}} &
\color{white}\shortstack{\textbf{Multi-}\\\textbf{vehicle}} &
\color{white}\shortstack{\textbf{Human-}\\\textbf{only}} &
\color{white}\textbf{Primary evaluation} &
\color{white}\textbf{Confound control} \\
\midrule

\rowcolor{DriveMint!38}
\multicolumn{9}{l}{\textbf{Naturalistic data resources}} \\
SHRP2 NDS~\cite{shrp2nds} & Real & 3{,}400+ drivers; 5.4M+ trips & Video, vehicle state, assessments & \cmark & \cmark & \cmark & Safety and behavior analysis & Study-specific analyses \\
HDD~\cite{hdd} & Real & 104 h; one instrumented vehicle & Video, CAN, GPS/IMU & \xmark & \xmark & \cmark & Maneuver and causal-scene understanding & \xmark \\
comma2k19~\cite{comma2k19} & Real & 33 h; 2{,}019 clips; fixed corridor & Video, CAN, GNSS/IMU & \xmark & \xmark & \pmark & Localization and mapping & \xmark \\
\addlinespace[1pt]

\rowcolor{DriveGold!32}
\multicolumn{9}{l}{\textbf{Driver identification and style studies}} \\
Driver2vec~\cite{driver2vec} & Controlled & 51 drivers; 10-s samples & Vehicle telemetry & \cmark & \xmark & \cmark & Closed-set driver identification & Scenario-consistency checks \\
Lyu et al.~\cite{lyu2022naturalistic} & Real & 44 drivers; two expressways & ADAS signals, video & \cmark & \xmark & \cmark & Three-class style recognition & Selected longitudinal conditions \\
\addlinespace[1pt]

\rowcolor{DriveBlue!22}
\multicolumn{9}{l}{\textbf{Personalized driving datasets and benchmarks}} \\
PDB~\cite{pdb} & Real & 12 drivers; 451 min; 6.6 TB & LiDAR, video, CAN, IMU, physiology & \cmark & \xmark & \cmark & Driver analysis and trajectory prediction & Same vehicle, route, and lighting \\
StyleDrive~\cite{styledrive} & Real & $\sim$30k preference-labeled scenes & Scene sensors and trajectories & \xmark & \pmark & \cmark & Style-conditioned E2E planning & Scenario-conditioned preference labels \\
PDB-Eval~\cite{pdbeval} & Real & PDB-X and PDB-QA & Multi-view video and language & \pmark & \xmark & \cmark & Behavior description and explanation & Inherits PDB collection control \\
Person2Drive~\cite{person2drive} & Sim. & Scalable personalized CARLA drives & Camera, ego state, trajectories & \cmark & \pmark & \cmark & Closed-loop personalized E2E driving & Simulation control; distribution metrics \\
\midrule
\rowcolor{DriveGreen!22}
\textbf{DriveDNA (ours)} & \textbf{Real} & \textbf{465 drivers; 115 models; 4{,}121 drives; 975 hours} & \textbf{Video, CAN, radar, lane/pose} & \cmark & \cmark & \cmark & \textbf{Re-ID, personalized prediction, condition-matched comparison} & \textbf{Matched driving conditions + vehicle/route/condition leakage probes} \\
\bottomrule
\end{tabularx}
\end{table*}

\textbf{Driver identification and driving-style modeling.}
Vehicle signals can reveal driver identity, from early CAN-based fingerprinting~\cite{enev2016automobile} and raw in-vehicle-network classification~\cite{remeli2019automatic} to learned embeddings such as Driver2vec~\cite{driver2vec}.
These studies establish that short telemetry sequences contain identity information, but most emphasize closed-set recognition in fixed or weakly controlled data configurations; vehicle dynamics, repeated routes, and context exposure can therefore remain correlated with identity without being measured separately.
Human-factors and intelligent-vehicle reviews instead characterize driving style as habitual behavior that is relatively stable across situations~\cite{sagberg2015review,chu2023review,tselentis2023driver,liao2025personalization}.
Chu et al.~\cite{chu2023review} distinguish short-term, scene-dependent behavior from long-term style, while naturalistic studies show both context-specific adaptation~\cite{wen2022carfollowing} and repeatable longitudinal differences~\cite{lyu2022naturalistic}.
Maneuver-specific personalized models, such as the Driver Digital Twin for lane-change prediction~\cite{liao2023ddt}, further demonstrate the predictive value of driver history but do not provide a general-purpose benchmark across behaviors and vehicles.
DriveDNA adopts these insights operationally: short-term behaviors define the annotation vocabulary, whereas stable style is evaluated through cross-drive re-identification, personalized future-motion prediction, and matched-context comparison, with shortcut leakage reported explicitly.

\textbf{Personalized driving benchmarks and planning methods.}
PDB~\cite{pdb} isolates individual differences through a small-scale same-vehicle, same-drive collection.
StyleDrive~\cite{styledrive} attaches preference labels to nearly 30{,}000 real-world scenes for style-conditioned end-to-end planning, PDB-Eval~\cite{pdbeval} evaluates multimodal description and explanation of personalized behavior, and Person2Drive~\cite{person2drive} provides scalable simulated data and closed-loop personalized evaluation.
Related planning methods address complementary modeling questions: DiffusionDrive~\cite{diffusiondrive} models multimodal end-to-end trajectories without persistent driver identities, while PLAN-S~\cite{plans} injects explicit style-conditioned cost maps into latent-world-model planners.
DriveDNA differs in objective: it transfers controlled driver comparison to naturalistic fleet data, replacing experimental or simulation control with matched-context evaluation, and evaluates representations and predictors rather than the planner alone.

\textbf{Representation learning under confounding.}
Modern time-series and multimodal models provide the components needed to test this benchmark, including patch- and variate-based Transformers~\cite{patchtst,itransformer}, contrastive and margin objectives~\cite{supcon,arcface}, prototypical few-shot evaluation~\cite{protonet}, self-supervised time-series pretraining~\cite{ts2vec}, time-series foundation models~\cite{moment}, frozen image and video encoders~\cite{dinov2,dinov3,siglip2,vjepa2}, conditional fusion~\cite{film,transfuser}, domain-adversarial learning~\cite{dann}, and vision--language annotation~\cite{qwen3vl}.
We use these as controlled baselines rather than methodological contributions; the contribution is a unified protocol that compares their driver utility, predictive value, and vehicle/route/context leakage under the same frozen splits.

\section{DriveDNA Dataset Development}
\label{sec:dataset}

DriveDNA is built from logs recorded during participants' everyday driving by windshield-mounted comma devices~\cite{comma_device} running the open-source openpilot logging stack~\cite{openpilot}, later contributed voluntarily to the study.
During human-driven intervals, each device acts as a dashcam-style logger, synchronizing front-view road video with CAN-derived kinematics, driver inputs, and available radar, lane, and pose estimates.
The raw dataset comprises 465 drivers, 115 distinct vehicle models among 26 vehicle brands, and 4{,}373 drives; drives with unidentified vehicle models are excluded only from the vehicle-model count.

\begin{figure}[htbp]
    \centering
    \includegraphics[width=0.75\columnwidth]{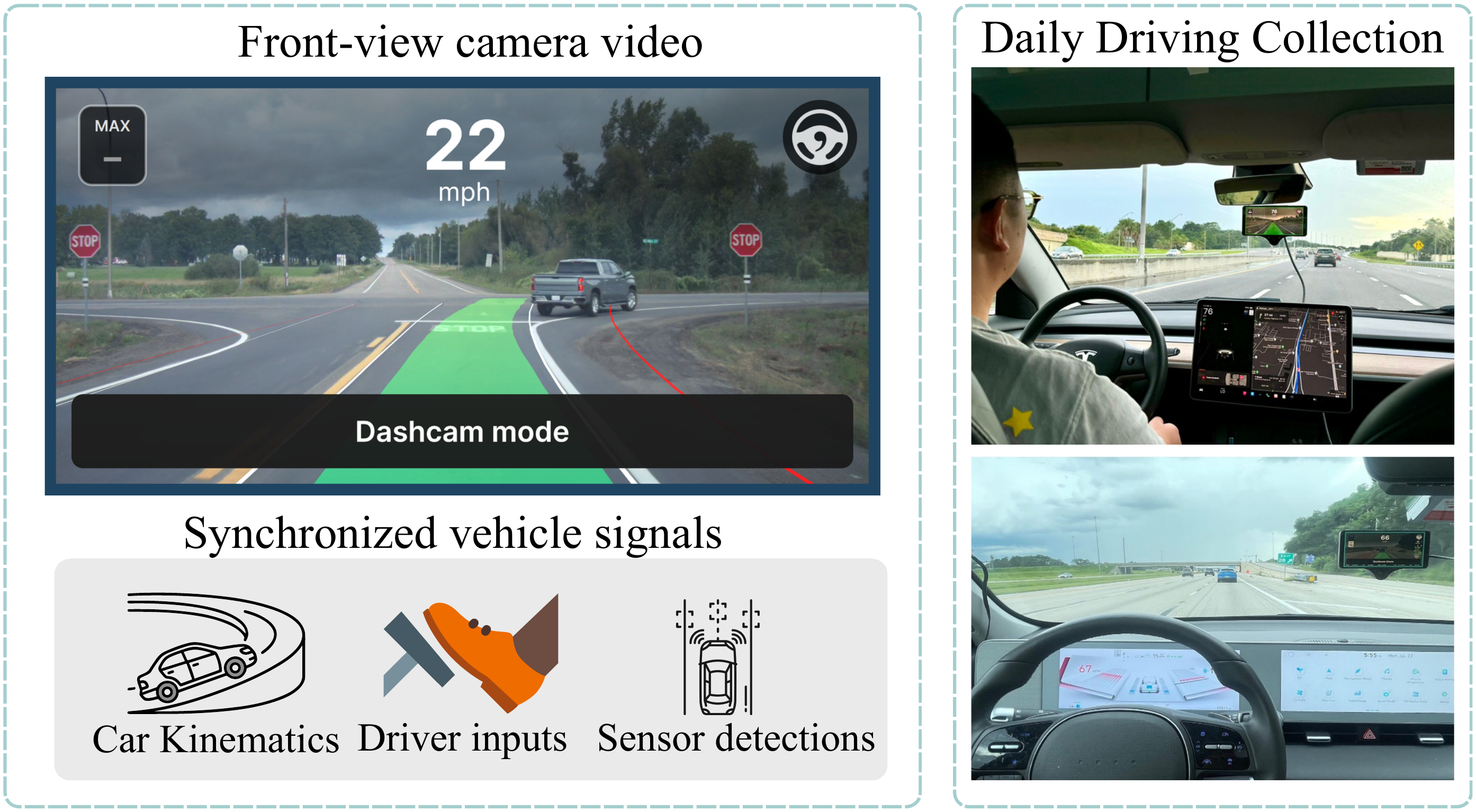}
    \caption{\textbf{DriveDNA data collection.}
    A windshield-mounted logger synchronously records front-view video and vehicle signals during routine driving.}
    \label{fig:data_collection}
\end{figure}

\begin{figure*}[htbp]
\centering
\includegraphics[
  page=2,
  width=0.8\textwidth
]{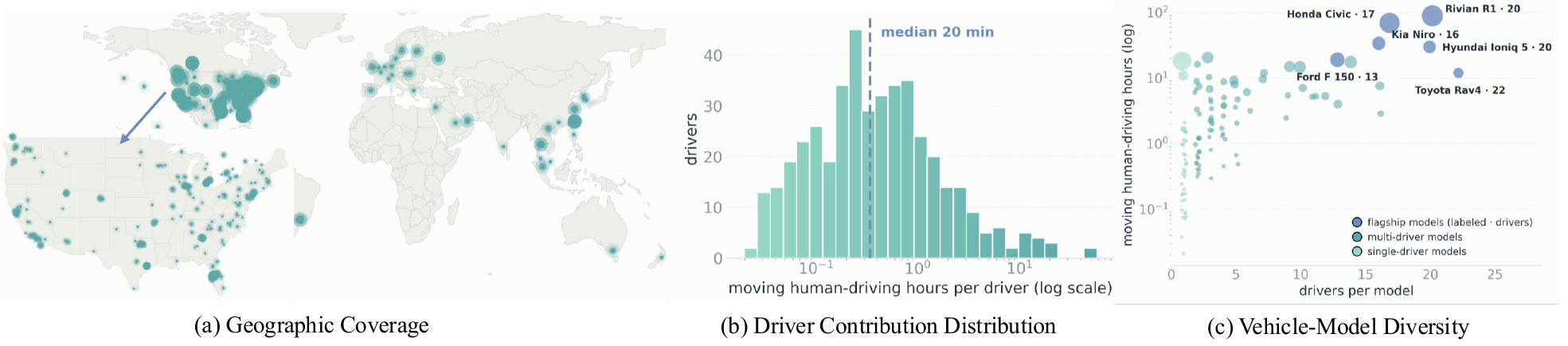}
\caption{
DriveDNA distribution.
\textbf{(a)} Geographic coverage of drives with GPS from March 2023 to July 2026.
\textbf{(b)} Driver contribution over 449 drivers with moving human-driving (HD) data.
\textbf{(c)} HD hours and driver coverage across 115 vehicle models.
}
\label{fig:dataset_distribution}
\end{figure*}

As summarized in Figure~\ref{fig:dataset_distribution}, the dataset is diverse along three dimensions.
The 938 drives with GPS metadata available for aggregate analysis cover multiple continents, with the highest density in North America and additional coverage across Europe, Asia, Africa, and Oceania. Collection extends from March 2023 to July 2026, capturing naturalistic contributions over more than three years rather than through a single controlled campaign.
Driver contribution is strongly heavy-tailed.
Of the 465 drivers in the corpus, 449 contribute at least some human-controlled driving with vehicle speed above \(2\,\mathrm{m/s}\); the remaining 16 drivers have only stopped or very-low-speed human-driving records under this threshold.
The median moving HD time is 20 minutes, while the ten most represented drivers account for 39\% of the total.
Vehicle coverage is similarly heterogeneous.
The most widely shared models are the Toyota RAV4 (22 drivers), Rivian R1 and Hyundai Ioniq 5 (20 drivers each), and Honda Civic (17 drivers); overall, 420 drivers with identified vehicle models share their model with at least one other driver, while a long tail of models is represented by a single driver.
These distributions position DriveDNA as a broad naturalistic benchmark spanning geographic regions, collection periods, driver contribution levels, and vehicle platforms.
Each drive is additionally associated with vehicle-specific physical parameters, including mass, wheelbase, and steering ratio, which support the vehicle-aware splits, matched comparisons, and leakage diagnostics in the benchmark. Table~\ref{tab:corpus} (Appendix~\ref{app:distribution}) shows the dataset statistics.

\subsection{Data Processing and HD Extraction}
\label{sec:dataset:preprocessing}

\textbf{Decoding and signal alignment.}
We decode all drives into a common 10\,Hz signal table.
To prevent models from exploiting differences between the two available log formats, every drive is decoded from the same format; this reduces format-prediction performance to chance (0.50 AUROC; Appendix~\ref{app:decode}).
We represent lateral motion using realized path curvature rather than steering-wheel angle, since steering ratio and wheelbase vary across vehicles.
Steering-wheel angle is retained as an auxiliary signal, while path curvature serves as the primary lateral-motion variable.
Section~\ref{sec:results} shows that steering angle contains more vehicle information than realized curvature.
Of the 4{,}373 raw drives, 4{,}121 are successfully decoded and included in the released dataset.

\textbf{Human-driving extraction.}
We remove all automation-active intervals identified from the vehicle logs, retaining only human-controlled driving.
Continuous human-driving segments are then extracted using a 1\,s persistence rule and a 2\,s minimum gap, with segments shorter than 30\,s discarded.
This yields 12{,}440 segments from 3{,}989 drives and 452 drivers, totaling 975 hours of human-controlled driving, including 581 hours at speeds above 2\,m/s.
All benchmark tasks use only these segments.

\subsection{Signal Groups}
\label{sec:dataset:signals}

Because signal availability differs across vehicle models, we organize the released variables into three groups by their role in the benchmark.
\textbf{Tier A: realized vehicle motion.}
This group includes speed, longitudinal acceleration, jerk, path curvature, yaw-rate-derived variables, lane offset, lead-vehicle distance, time headway, and time-to-contact.
These widely available signals form the primary inputs, prediction targets, and comparison variables.
\textbf{Tier B: driver inputs.}
This group includes gas, brake, steering-wheel angle, and steering-wheel rate.
Coverage is vehicle-dependent: gas is available in 40\% of drives, brake in 57\%, and steering signals in 95\%.
These signals are therefore treated as auxiliary inputs rather than required benchmark variables.
\textbf{Tier C: behavioral primitives.}
This group contains derived indicators such as close following, large headway, hard brake, high jerk, sharp steering, lane correction, and curve-entry deceleration.
They support behavioral annotation, condition matching, and stratified evaluation.

\subsection{Windows and Behavioral Labels}
\label{sec:dataset:windows}

We divide the moving parts of each HD segment into 60\,s windows with a 30\,s stride.
This produces 62{,}674 windows from 428 of the 465 drivers; 355 of them have sufficient data to enter the frozen train/validation/test/few-shot folds (Appendix~\ref{app:splits}).
Each window is assigned one of six driving contexts: car-following, free driving, curve, stop-and-go, high-speed, or urban driving.
The labels are based on speed, curvature, lead-vehicle information, and stopping patterns.
They describe the driving situation rather than the driver's style.
Each window is also tagged with eight behavioral primitives.
For each primitive, a window is labeled high when its value is above the 80th percentile within the same scenario, and low when it is below the 20th percentile.
Using context-specific thresholds reduces the effect of road type and speed differences.
These primitives are weak labels for short-term behavior, not fixed personality or driving-style labels.
A human audit of 240 windows reaches 93.0\% agreement, and the results remain unchanged when Q75 or Q85 thresholds are used (Appendix~\ref{app:annotation}).

\subsection{Maneuver-Event Annotations}
\label{sec:dataset:maneuvers}

DriveDNA includes 276{,}248 maneuver-event annotations across six classes: deceleration, acceleration, intersection turn, curve driving, car-following state, and lane change.
These events are detected using rule-based methods based on vehicle motion, steering, lane, and lead-vehicle signals.
We assess annotation quality by reviewing the corresponding forward-video clips. For the first five classes, approximately 1{,}000 events per class were randomly sampled, with precision ranging from 94.8\% to 99.6\%. 
 Lane changes are harder to detect automatically because lane-signal quality and road geometry vary across vehicles and environments: an individual review of all 36{,}488 watchable candidates yields a candidate precision of 61.2\%, and only the 22{,}322 verified lane changes are released as events, with the rejected candidates retained as labeled hard negatives. All detected events are retained for corpus-level analysis and behavioral statistics, but are not used as training targets or ground-truth labels in the core benchmark. Detailed detection rules, event counts, and audit results are provided in Appendix~\ref{app:maneuvers}.

\subsection{Release Considerations}
\label{sec:dataset:privacy}

DriveDNA is released in two levels.
(i) De-identified 10\,Hz vehicle signals, frozen video embeddings, data splits, and code are publicly available.
(ii) Raw forward video is available only under a data-use agreement, with faces and license plates blurred.
Driver identifiers are salted-hashed, and VINs, origin driver-id, and GPS coordinates are removed from the released data.
Section~\ref{sec:release} describes the release policy, governance, and intended uses thoroughly.

\section{Benchmark Design}
\label{sec:benchmark}

DriveDNA poses three questions:
does a stable driver signature exist (\emph{identity}),
does driver information improve future-motion prediction (\emph{utility}),
and do driver-associated differences remain under matched driving conditions (\emph{robustness})?
These define three core tasks, evaluated with fixed metrics, splits, and procedures released with the benchmark harness (Section~\ref{sec:release}).

\subsection{Benchmark Tasks}
\label{sec:benchmark:tasks}

The benchmark comprises three core tasks, behavioral labels for stratified analysis, and two optional extensions (Table~\ref{tab:tasks}).

\textbf{Driver re-identification and verification.}
Given $k \in \{1,3,5,10\}$ minutes of a driver's windows as support, identify or verify that driver among unseen drivers, using query windows from different drives.
We report top-$k$ accuracy, AUROC, and EER following speaker-recognition enrollment--verification protocols~\cite{heigold2016endtoend,huh2024voxsrc}.

\textbf{Personalized behavior prediction.}
From a 5\,s history at 10\,Hz, predict acceleration and path curvature over 1, 3, and 5\,s horizons; 3\,s is the primary horizon.
Models may use front-video scene information, vehicle parameters, and a few-shot driver support set.
Each personalized model is compared against a non-personalized counterpart with the same architecture.

\textbf{Condition-matched comparison.}
This task uses 14{,}868 balanced window pairs, matched on vehicle model, scenario type, speed range, and headway range where applicable.
The model predicts whether two matched windows come from the same driver, bringing the controlled comparisons of laboratory style studies into naturalistic data.
Matching quality is independently validated with vision--language scene attributes (Appendix~\ref{app:vlm}).

\textbf{Behavioral labels.}
The eight weak labels of Section~\ref{sec:dataset:windows} provide an interpretable behavior vocabulary, the stratification variable for prediction, and the variables for condition-matched comparison.

\textbf{Event forecasting and style explanation.}
Event forecasting asks whether a hard-braking or sharp-steering event begins within the next 1--5\,s, reported as AP and AUROC by lead time.
Style explanation produces a post-hoc event category and an explanation grounded in visual and CAN evidence; we treat it as exploratory.
The three predictive tasks are disjoint: a future sequence, a future event label, and a post-hoc account, respectively.

\textbf{The role of video.}
Front-view video enters the benchmark as contextual input only: as context tokens for the forecasting tasks, as an independent check on condition-matched pairs, and as evidence for style explanation.
Video frames carry strong route and scene information that can act as a shortcut for driver re-identification (Section~\ref{sec:results}), so video-based representations are always evaluated together with the route- and context-leakage probes.

\begin{table}[htbp]
\centering
\caption{The DriveDNA task suite.}
\label{tab:tasks}
\setlength{\tabcolsep}{3.6pt}
\renewcommand{\arraystretch}{1.00}
\footnotesize

\begin{tabularx}{\linewidth}{
  >{\raggedright\arraybackslash}p{0.25\linewidth}
  >{\raggedright\arraybackslash}X
  >{\raggedright\arraybackslash}p{0.25\linewidth}
}
\toprule
\rowcolor{DriveTeal}
\color{white}\textbf{Task} &
\color{white}\textbf{Input $\rightarrow$ Output} &
\color{white}\textbf{Metrics} \\
\midrule

\rowcolor{DriveMint!38}
\multicolumn{3}{l}{\textbf{Core tasks}} \\

Driver re-identification &
$k$-min support $\rightarrow$ driver identity &
Top-$k$, AUROC, EER \\

Personalized behavior prediction &
5\,s history $\rightarrow$ 1--5\,s future motion &
RMSE, PG, MMD/KL/W1 \\

\rowcolor{DriveGreen!12}
Condition-matched comparison &
Matched window pair $\rightarrow$ same driver? &
AUROC, EER \\

\addlinespace[1pt]
\rowcolor{DriveGold!32}
\multicolumn{3}{l}{\textbf{Supporting annotation}} \\

Behavioral primitives &
60\,s window $\rightarrow$ 8 weak labels &
Audit agreement \\

\addlinespace[1pt]
\rowcolor{DriveBlue!22}
\multicolumn{3}{l}{\textbf{Optional tasks}} \\

Event forecasting &
5\,s history $\rightarrow$ event in 1--5\,s &
AP, AUROC, lead time \\

Style explanation &
Event window $\rightarrow$ category + evidence &
Category accuracy (exploratory) \\

\bottomrule
\end{tabularx}
\end{table}

\subsection{Benchmark Splits}
\label{sec:benchmark:splits}

\begin{figure}[htbp]
\centering
\includegraphics[
  page=3,
  width=0.8\linewidth
]{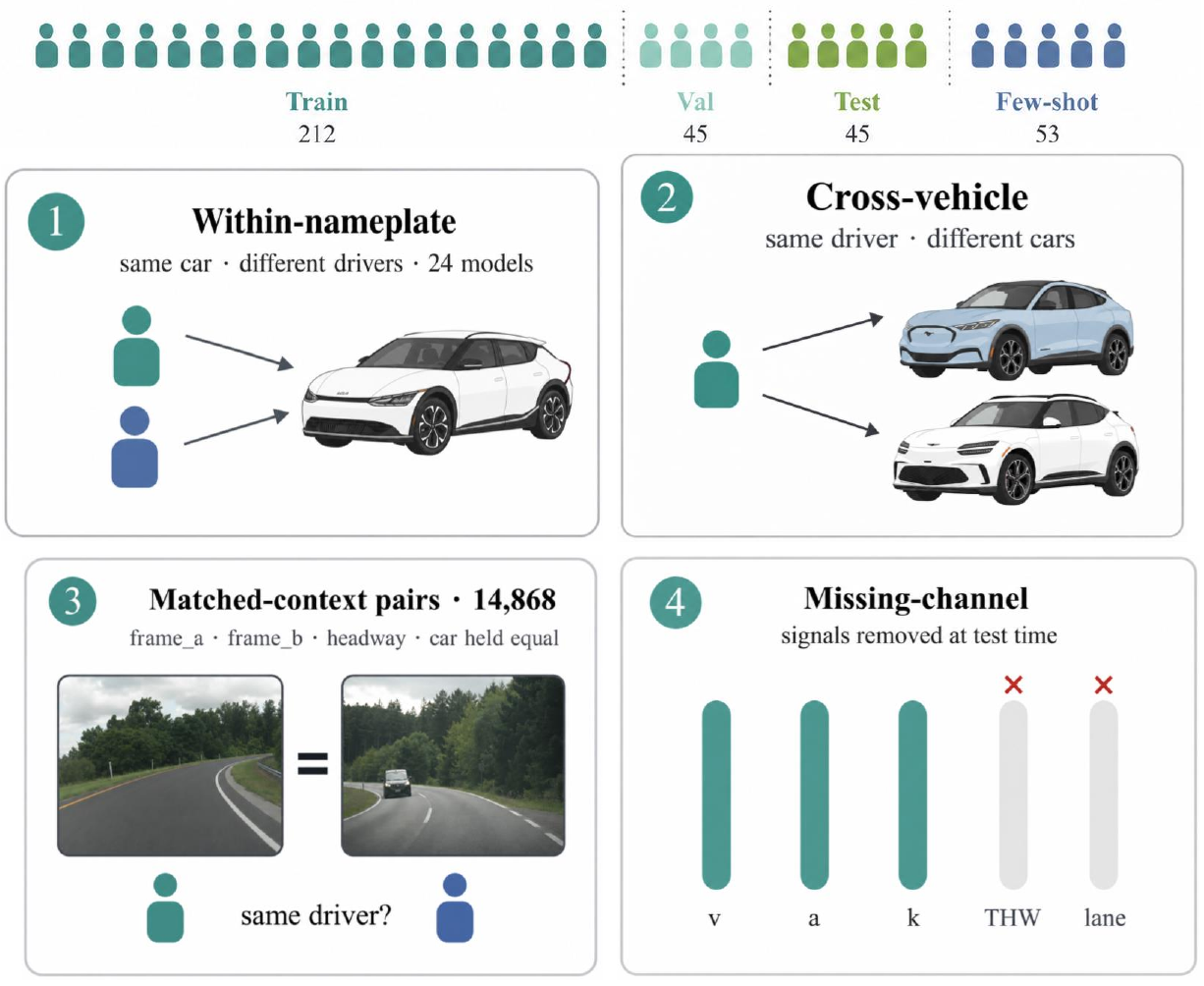}
\caption{
DriveDNA is driver-disjoint, with evaluations:
(1) different drivers in same vehicle model,
(2) same driver across vehicles,
(3) matched driving contexts and
(4) missing signals.
}
\label{fig:benchmark_splits}
\end{figure}

Figure~\ref{fig:benchmark_splits} summarizes the benchmark splits.
The driver-disjoint split contains 212 training, 45 validation, and 45 test drivers (70\%/15\%/15\%), with a separate 53-driver hold-out for few-shot evaluation.
The hold-out drivers appear in no training fold and are used only for unseen-driver enrollment; support and query windows always come from different drives.

Additional manifests isolate specific sources of generalization: within-nameplate evaluation compares different drivers of the same vehicle model across 24 models; a cross-vehicle split covers drivers observed in more than one vehicle; the condition-matched manifest fixes driving condition; and missing-channel tests remove signal groups at evaluation time.

\subsection{Evaluation Metrics}
\label{sec:benchmark:metrics}

Driving style is reflected not only in driver recognition, but also in future behavior and behavior distributions.
The evaluation harness reports five groups of metrics.

(i) Recognition metrics across the enrollment curve: top-$k$ accuracy, AUROC, and EER.

(ii) Prediction error at each horizon: sequence RMSE over the predicted acceleration and curvature series, plus final-step error at 1, 3, and 5\,s.

(iii) Distribution distances between predicted and observed behavior: MMD, KL divergence, and Wasserstein-1.
These distances are computed within driving-scenario groups so that differences in scenario mix are not mistaken for driver style.
Distribution estimation and aggregation are fixed in the released evaluation harness; Appendix~\ref{app:baselines} provides the exact settings.

(iv) Personalization gain: $\mathrm{PG} = (E_{\mathrm{generic}}-E_{\mathrm{personalized}})/E_{\mathrm{generic}} \times 100\%$ for error-based metrics, and $\mathrm{PG}_{\mathrm{NLL}} = \mathrm{NLL}_{\mathrm{generic}} - \mathrm{NLL}_{\mathrm{personalized}}$ (in nats) for likelihood; positive values indicate a benefit from personalization.

(v) Leakage scores, which measure how well vehicle model, drive identity, and driving condition can be predicted from a learned representation, a direct measurement of shortcut learning~\cite{geirhos2020shortcut}. 
We summarize driver utility and shortcut leakage using utility--leakage Pareto plots.

\subsection{Evaluation Protocol}
\label{sec:benchmark:protocol}

Two rules from our evaluation audit are part of the benchmark protocol.
First, evaluation manifests are fixed across runs, and random seeds affect training only: per-window likelihood values are heavy-tailed, and letting the evaluated samples move with the seed can inflate measured personalization effect by an order of magnitude (Appendix~\ref{app:protocol}).
Second, every headline result is reported as the mean and standard deviation over three training seeds under a fixed evaluation split.
All results in Section~\ref{sec:results} follow these rules.

\section{Baseline Settings}
\label{sec:baselines}

All baselines train and evaluate on the frozen splits.
We group thirty baseline configurations into five modeling dimensions aligned with the benchmark questions: representation learning, shortcut robustness, personalization, multimodal modeling, and distributional prediction.
Together, they provide controlled reference points for measuring driver identity, predictive utility, and robustness under condition matching.
Details and hyperparameters are in Appendix~\ref{app:baselines}.

\textbf{Representation.}
We evaluate driver representations on unseen drivers.
We train patch-based Transformer encoders (joint-channel and channel-independent PatchTST~\cite{patchtst}, iTransformer~\cite{itransformer}) with supervised-contrastive \cite{supcon} or ArcFace~\cite{arcface} objectives, and evaluate them with prototype-based few-shot enrollment~\cite{protonet}.
A frozen time-series foundation model (MOMENT~\cite{moment}) gives a zero-shot reference row.
Two self-supervised rows (masked reconstruction, and a JEPA-style latent-predictive baseline) measure how far label-free pretraining closes the gap.

\textbf{Shortcut control.}
We measure driver-specific information after controlling for vehicle and driving scenario.
A population model conditioned on state, context, and vehicle gives ResidualStyle: each driver is represented by what the population model cannot explain.
Adversarial invariance (DANN~\cite{dann}) and the utility--leakage Pareto curve complete this stage.

\textbf{Personalization.}
We compare driver information mechanisms for future-motion prediction.
We build a conditioning ladder from a generic model to few-shot support encoding, injected through FiLM~\cite{film}.
A query-injection variant (in the style of TransFuser~\cite{transfuser}) tests whether the conditioning architecture itself matters.

\textbf{Multimodal context.}
We compare frame-level and temporal video features for context modeling.
The main baseline (multimodal context-conditioned predictor, MCPP) fuses CAN tokens with frozen video tokens through cross-attention, under a six-way conditioning ablation: CAN, +vehicle, +video, +driver, +video+driver, +residual.
The video encoder is its own ablation axis: per-frame features from three generations (DINOv2~\cite{dinov2}, DINOv3~\cite{dinov3}, SigLIP2~\cite{siglip2}) against temporal clip features (V-JEPA~2~\cite{vjepa2}).
This contrasts appearance features with motion features.
A CLIP-style CAN$\leftrightarrow$video contrastive baseline and a vision--language attribute extractor (Qwen3-VL~\cite{qwen3vl}) cover multimodal self-supervision and semantic context.

\textbf{Distributional prediction.}
We model behavior uncertainty and multiple future modes.
Mixture-density and conditional-VAE heads on the behavior predictor, evaluated with likelihood and distribution metrics of Section~\ref{sec:benchmark:metrics}.

Throughout, interpretable descriptors of the CAN, radar, and lane signals, paired with gradient-boosted classifiers, serve as a lower-bound anchor and as an interpretation probe.
They are classical on purpose and are not a benchmark task.
Diagnostic probes (video-only re-identification; vehicle, route, and context leakage) run alongside every stage.
Table~\ref{tab:baseline_coverage} (in Appendix~\ref{app:protocol}) summarizes the coverage.
Appendix~\ref{app:vintage} lists every component and details.

\section{Results}
\label{sec:results}

DriveDNA shows that driver-specific information remains under vehicle and driving-condition controls, but strong re-identification does not prove driving style or predictive utility.
We first quantify these sources of variation and compare learned representations, then test them through condition-matched comparison, personalized prediction, and route- and vehicle-leakage analyses.
All results follow the fixed protocol in Section~\ref{sec:benchmark:protocol}; headline values are reported as mean $\pm$ standard deviation over three training seeds, with driver-level bootstrap intervals where applicable.

\subsection{Driver-Specific Signal Under Confounding}
\label{sec:results:confound}

We first observe a substantial portion of apparent between-driver variation attributable to driving conditions and vehicle characteristics.
Population models conditioned on scenario, speed regime, and vehicle explain 29--60\% of the variance of naive behavior statistics.
Yet a stable driver-specific signal survives: after residualizing against these models, drivers remain identifiable at $4\times$ chance.
Signal choice also matters: vehicle-model probes reach $2.3\times$ chance from steering-wheel angle but chance from realized curvature, indicating that evaluations without vehicle and driving-condition controls may overstate driver-specific signal.

\subsection{Representation Comparison}
\label{sec:results:ladder}

Table~\ref{tab:t2} shows the re-identification comparison on unseen drivers.
Classical descriptors reach AUROC .707.
A supervised contrastive Transformer encoder reaches $.935 \pm .005$, with top-1 identification at about $33\times$ chance.
The gaps between architectures are three to six times the seed noise: channel-independent PatchTST matches the joint encoder ($.932$ vs.\ $.935$), ArcFace trails SupCon ($.902$), and iTransformer trails both ($.877$).
Label-free pretraining closes most of the supervised gap (masked reconstruction $.907$; the JEPA-style latent-predictive baseline $.878$).
Frozen foundation models land at or below the descriptor level: the MOMENT-1 time-series model reads $.636$ zero-shot, and decoder LLMs embedding the same windows as serialized text (Qwen3-4B $.596$, Llama-3.2-3B $.598$) do no better---generic pretraining, whether over time series or language, does not encode driver identity without adaptation (Appendix~\ref{app:llm}).

\begin{table}[htbp]
\centering
\caption{
Driver re-identification with 5-minute enrollment on unseen drivers.
}
\label{tab:t2}
\setlength{\tabcolsep}{3.0pt}
\renewcommand{\arraystretch}{0.98}
\footnotesize

\begin{tabularx}{\linewidth}{
  >{\raggedright\arraybackslash}X
  >{\centering\arraybackslash}p{0.18\linewidth}
  >{\centering\arraybackslash}p{0.13\linewidth}
  >{\centering\arraybackslash}p{0.13\linewidth}
}
\toprule
\rowcolor{DriveTeal}
\color{white}\textbf{Representation} &
\color{white}\textbf{AUROC $\uparrow$} &
\color{white}\textbf{EER $\downarrow$} &
\color{white}\textbf{Top-1 $\uparrow$} \\
\midrule

\rowcolor{DriveMint!16}
\multicolumn{4}{l}{\textbf{Reference and zero-shot baselines}} \\

Descriptors (anchor) &
.707 & .349 & .09 \\

Qwen3-4B text encoder (zero-shot) &
.596 & .431 & .05 \\

Llama-3.2-3B text encoder (zero-shot) &
.598 & .428 & .05 \\

MOMENT-1 (zero-shot) &
.636 & .408 & .08 \\

\addlinespace[1pt]
\rowcolor{DriveBlue!12}
\multicolumn{4}{l}{\textbf{Label-free representation learning}} \\

CLIP-aligned CAN (no labels) &
.831 & .244 & .21 \\

JEPA-style SSL + probe &
.878 & .197 & .20 \\

Masked-TS SSL + probe &
.907 & .166 & .29 \\

\addlinespace[1pt]
\rowcolor{DriveGold!14}
\multicolumn{4}{l}{\textbf{Supervised representation learning}} \\

iTransformer + SupCon &
.877 $\pm$ .008 & .192 & .23 \\

ArcFace &
.902 $\pm$ .005 & .174 & .39 \\

PatchTST-CI + SupCon &
.932 $\pm$ .002 & \textbf{.127} & .32 \\

\rowcolor{DriveGreen!12}
\textbf{PatchTST + SupCon} &
\textbf{.935 $\pm$ .005} &
\textbf{.127} &
\textbf{.44} \\

\bottomrule
\end{tabularx}
\end{table}

\subsection{Condition-Matched Evaluation}
\label{sec:results:matched}

Table~\ref{tab:t3full} evaluates every representation on the 14{,}868 matched-context pairs.
Descriptor verification drops to AUROC .550---chance level---while the learned embedding holds at $.811 \pm .006$, consistently across all six scenario types (.78--.83).
Three patterns emerge.
First, the ranking of trained CAN encoders is preserved under matching (SupCon and ArcFace on top, iTransformer lowest), suggesting the task measures representation quality rather than reshuffling it.
Second, every zero-shot foundation model falls to chance or near it (.55--.60), consistent with their unmatched scores being carried largely by context.
Third, the two representations with visual grounding drop the most: the video-only probe falls from .937 to .675---the largest drop of any representation---and the CLIP-aligned CAN encoder, trained to agree with video, shows a similarly outsized drop (.831 $\rightarrow$ .683, below the label-free SSL encoders it beats without matching), consistent with alignment to video importing context dependence.
The vision--language validation of Section~\ref{app:vlm} confirms that the matching itself is visually credible.
The condition-matched task therefore separates driver-specific behavior from exposure to similar circumstances, and it does so for every class of representation at once.
Matching holds the vehicle model rather than the physical vehicle fixed; Appendix~\ref{app:vehinstance} evaluates residual vehicle effects through cross-vehicle verification.

\begin{table}[htbp]
\centering
\caption{Representation AUROC before and after condition matching on unseen drivers.}
\label{tab:t3full}
\setlength{\tabcolsep}{3.0pt}
\renewcommand{\arraystretch}{1.02}
\footnotesize

\begin{tabularx}{\linewidth}{
  >{\raggedright\arraybackslash}X
  >{\centering\arraybackslash}p{0.17\linewidth}
  >{\centering\arraybackslash}p{0.20\linewidth}
  >{\centering\arraybackslash}p{0.14\linewidth}
}
\toprule
\rowcolor{DriveTeal}
\color{white}\textbf{Representation} &
\color{white}\textbf{Unmatched} &
\color{white}\textbf{Matched} &
\color{white}\textbf{$\Delta$} \\
\midrule

\rowcolor{DriveMint!16}
\multicolumn{4}{l}{\textbf{Reference and zero-shot baselines}} \\

Descriptors (anchor) &
.707 & .550 & $-.157$ \\

Qwen3-4B text (zs) &
.596 & .551 & $-.045$ \\

MOMENT-1 (zs) &
.636 & .596 & $-.040$ \\

\addlinespace[1pt]
\rowcolor{DriveBlue!12}
\multicolumn{4}{l}{\textbf{Multimodal and label-free representations}} \\

\rowcolor{DriveGold!10}
Video-only probe &
.937 & .675 & \textbf{$-.262$} \\

CLIP-aligned CAN &
.831 & .683 & $-.148$ \\

JEPA-style SSL &
.878 & .735 & $-.143$ \\

Masked-TS SSL &
.907 & .740 & $-.167$ \\

\addlinespace[1pt]
\rowcolor{DriveGold!14}
\multicolumn{4}{l}{\textbf{Supervised representation learning}} \\

iTransformer + SupCon &
.877 & .762 $\pm$ .001 & $-.115$ \\

PatchTST-CI + SupCon &
.932 & .777 $\pm$ .004 & $-.155$ \\

ArcFace &
.902 & .807 $\pm$ .006 & $-.095$ \\

\rowcolor{DriveGreen!12}
\textbf{PatchTST + SupCon} &
.935 &
\textbf{.811 $\pm$ .006} &
$-.124$ \\

\bottomrule
\end{tabularx}
\end{table}

\subsection{Personalization Gains}
\label{sec:results:pg}

Having established that a driver-specific signal survives condition matching, we next test whether it translates into predictive utility.
Few-shot personalization improves prediction on unseen drivers in all three seeds, with the gain concentrated at short horizons (positive at 1--5\,s; $+0.8 \pm 0.5\%$ at the 5\,s horizon, $k{=}5$).
The gain is small in RMSE but consistent in likelihood.
Both distributional heads improve when driver information is added, in every seed (CVAE $+6.9 \pm 4.3$ nats; MDN $+0.10 \pm 0.05$ nats), while their RMSE stays flat.
Adding driver information changes the predicted distribution rather than the point estimate; RMSE-only evaluation would not detect this effect.
Per-driver breakdowns support this reading: across three seeds, the likelihood gain is positive for 55\% of unseen drivers while the point-error gain is centered near zero, and personalization also moves the predicted behavior distributions closer to the observed ones under MMD and Wasserstein-1 (Appendix~\ref{app:distmetrics}).

Two negative results indicate that the evaluation procedure separates design choices rather than rewarding any use of driver information.
First, the best re-identification embedding, used as a conditioning signal, yields no prediction gain ($-0.2\%$); re-identification and prediction measure different capabilities.
Second, the same few-shot style vector that helps through FiLM conditioning ($+0.4$ to $+1.4\%$) hurts through query injection ($-0.7\%$): the conditioning architecture matters.

\subsection{Video: Identity Leakage and Context Value}
\label{sec:results:video}

A video-only probe re-identifies unseen drivers at AUROC .937, comparable to CAN.
The leakage probes suggest why: the video representation predicts the route at $347\times$ chance (CAN: $197\times$) and the vehicle at $64\times$ (CAN: $56\times$).
Controlling the vehicle separates the two channels.
On within-nameplate splits (same vehicle model, different drivers), CAN drops to .887---it loses the vehicle-dynamics cue---while video \emph{rises} to .962, because each driver still drives in their own places.
Matched-context evaluation adds a complementary control: when the compared windows share scenario, speed regime, and road context, the video probe drops from .937 to .675 (Table~\ref{tab:t3full})---the largest drop of any representation---consistent with much of its accuracy reflecting the shared context that matching removes.
These results are consistent with place recognition rather than driving behavior, which is why DriveDNA treats video as context.

We next examine which video features add predictive value beyond context.
Table~\ref{tab:video} in Appendix~\ref{app:llm} compares per-frame appearance features from three encoder generations against temporal clip features.
Per-frame features reduce event-forecasting AUROC in all three generations; DINOv3 shows the largest reduction (7.3 points), larger than its 2023 predecessor.
Per-frame features also add almost nothing to personalized prediction ($+0.09\%$).
Temporal clip features improve event forecasting (.842 $\rightarrow$ .847, monotonically in context length) and raise the video contribution in the six-way ablation to $+1.49 \pm 0.32\%$, with the full conditioning stack at $+2.1\%$.
Zero-shot vision--language and language models show the same pattern from a different angle: on a fixed anchor subsample they reach at best AUROC .759 against .857 for the trained CAN GRU, and forecasting ability does not track model recency (Appendix~\ref{app:llm}).

\subsection{Vehicle Effects}
\label{sec:results:vehicle}

The learned realized-motion embedding is close to vehicle-agnostic without any adversarial training: vehicle leakage sits at $2.6\times$ chance over 60+ models, and adversarial invariance (DANN) cannot reduce it further while it costs utility.
The residual $2.6\times$ reflects the natural entanglement of this fleet---each driver mostly owns one car.
An explicit learned vehicle embedding reduces prediction accuracy for unseen drivers in every configuration we tested ($-0.9$ to $-1.4\%$ across two training recipes and three seeds); the realized-motion channels appear to carry the vehicle implicitly, and an identity-style vehicle vector overfits the training fleet.
In this corpus, choosing vehicle-normalized targets at the data level (Section~\ref{sec:dataset:preprocessing}) contributed more to vehicle robustness than the invariance methods applied.

\section{Release, Privacy, and Ethics}
\label{sec:release}

\textbf{Tiered release.}
DriveDNA is released in two tiers.
The public tier contains the de-identified 10\,Hz signal tables, the frozen video embeddings (DINOv2, DINOv3, SigLIP2, V-JEPA~2), all split manifests, the VLM scene attributes, the evaluation harness, and the training code for every baseline in this paper.
The gated tier contains the raw forward video, with faces and license plates blurred, and requires a data-use agreement that restricts use to research.
The public tier alone reproduces every number in this paper.

\textbf{De-identification.}
To protect private information, driver identifiers are salted hashes, and the salt is not released.
VINs and device identifiers are further removed.
The released signals contain no GPS coordinates, and route identifiers are opaque strings.
We do not release cabin video or audio.

\textbf{Consent and governance.}
The data is collected from community drivers with informed consent for research use, and participants are compensated for contributing their data.
Collection followed the source platform's terms of use, which permit research use.
We document the collection terms, the de-identification pipeline, and the residual re-identification risks in the datasheet (supplementary material), together with a Croissant metadata record.
A takedown contact allows any driver to request removal of their data from future versions.

\textbf{Intended use and misuse.}
DriveDNA is intended for research on driving-style representation, personalized prediction, and evaluation methodology.
Section~\ref{sec:results:video} itself illustrates the main misuse risk: representations trained on naturalistic data can identify people through their routines rather than their behavior.
We therefore release leakage probes as part of the harness, ask users to report leakage alongside utility, and prohibit re-identification attempts in the data-use agreement.
However, insurance, employment, and law-enforcement scoring of individuals are out of scope of this study.

\textbf{Hosting.}
The dataset is hosted on Hugging Face with versioned releases and a DOI; the configuration and code tier is already public.

\section{Discussions and Conclusion}
\label{sec:conclusion}

\subsection{Discussion and Broader Impact}
The benchmark's central lesson is a three-way dissociation: driver identity, driving style, and personalized prediction are related but not interchangeable.
DriveDNA operationalizes this as a measurement checklist, leakage alongside utility, matched alongside unmatched performance, distributional alongside point metrics, so future style claims carry the controls needed to interpret them.
Beyond driving research, the corpus serves as a testbed for shortcut representation learning, gives driver-assistance developers pretraining data and a deployment-relevant test of when driver information helps, and offers privacy discussions a quantitative measure of how much identity and location vehicle telemetry can expose.

\subsection{Limitations}
The behavioral primitives are weak labels from percentile rules, not ground-truth style; we quantify their stability under threshold changes, but they should not be read as personality measurements.
The fleet skews toward one region and toward drivers who install an aftermarket driving system, and is not representative of all drivers.
Driver and vehicle are naturally entangled in this fleet---most drivers appear with one car---which bounds how far vehicle and driver effects separate; within-nameplate splits mitigate but do not remove this.
Pedal-level actuation is available on only part of the fleet, so realized motion, not actuation, defines the primary targets.
The event taxonomy is rule-derived; the release adds a six-class maneuver-event layer whose rules were human-audited at scale (Appendix~\ref{app:maneuvers}), with lane changes reviewed individually.
Finally, our audit shows that likelihood-based metrics are sensitive to the evaluation split; we fix the procedure, re-run everything, and document the process (Appendix~\ref{app:protocol}).

\subsection{Conclusion}
DriveDNA provides a large-scale, multi-vehicle, human-only naturalistic corpus for driving-style research, with a benchmark that treats confounding as something to measure rather than ignore; across thirty baselines it discriminates representations, personalization mechanisms, and evaluation procedures, and shows that apparent ``style'' is often vehicle, place, or driving conditions unless the evaluation accounts for them.

\clearpage
\bibliographystyle{ACM-Reference-Format}
\bibliography{references}


\begin{thebibliography}{46}

%%% the Web address.

\ifx \showCODEN    \undefined \def \showCODEN     #1{\unskip}     \fi
\ifx \showISBNx    \undefined \def \showISBNx     #1{\unskip}     \fi
\ifx \showISBNxiii \undefined \def \showISBNxiii  #1{\unskip}     \fi
\ifx \showISSN     \undefined \def \showISSN      #1{\unskip}     \fi
\ifx \showLCCN     \undefined \def \showLCCN      #1{\unskip}     \fi
\ifx \shownote     \undefined \def \shownote      #1{#1}          \fi
\ifx \showarticletitle \undefined \def \showarticletitle #1{#1}   \fi
\ifx \showURL      \undefined \def \showURL       {\relax}        \fi
% invisible to TeX
\providecommand\bibfield[2]{#2}
\providecommand\bibinfo[2]{#2}
\providecommand\natexlab[1]{#1}
\providecommand\showeprint[2][]{arXiv:#2}

\bibitem[Assran et~al\mbox{.}(2025)]%
        {vjepa2}
\bibfield{author}{\bibinfo{person}{Mido Assran}, \bibinfo{person}{Adrien
  Bardes}, \bibinfo{person}{David Fan}, \bibinfo{person}{Quentin Garrido},
  \bibinfo{person}{Russell Howes}, \bibinfo{person}{Matthew Muckley},
  \bibinfo{person}{Ammar Rizvi}, \bibinfo{person}{Claire Roberts},
  \bibinfo{person}{Koustuv Sinha}, \bibinfo{person}{Artem Zholus},
  {et~al\mbox{.}}} \bibinfo{year}{2025}\natexlab{}.
\newblock \showarticletitle{V-jepa 2: Self-supervised video models enable
  understanding, prediction and planning}.
\newblock \bibinfo{journal}{\emph{arXiv preprint arXiv:2506.09985}}
  (\bibinfo{year}{2025}).
\newblock


\bibitem[Bai et~al\mbox{.}(2025)]%
        {qwen3vl}
\bibfield{author}{\bibinfo{person}{Shuai Bai}, \bibinfo{person}{Yuxuan Cai},
  \bibinfo{person}{Ruizhe Chen}, \bibinfo{person}{Keqin Chen},
  \bibinfo{person}{Xionghui Chen}, {and} \bibinfo{person}{Zesen Cheng}.}
  \bibinfo{year}{2025}\natexlab{}.
\newblock \showarticletitle{Qwen3-VL Technical Report}.
\newblock \bibinfo{journal}{\emph{arXiv preprint arXiv:2511.21631}}
  (\bibinfo{year}{2025}).
\newblock


\bibitem[Caesar et~al\mbox{.}(2019)]%
        {nuscenes}
\bibfield{author}{\bibinfo{person}{Holger Caesar}, \bibinfo{person}{Varun
  Bankiti}, \bibinfo{person}{Alex~H. Lang}, \bibinfo{person}{Sourabh Vora},
  \bibinfo{person}{Venice~Erin Liong}, {and} \bibinfo{person}{Qiang Xu}.}
  \bibinfo{year}{2019}\natexlab{}.
\newblock \showarticletitle{nuScenes: A multimodal dataset for autonomous
  driving}.
\newblock \bibinfo{journal}{\emph{arXiv preprint arXiv:1903.11027}}
  (\bibinfo{year}{2019}).
\newblock


\bibitem[Chitta et~al\mbox{.}(2022)]%
        {transfuser}
\bibfield{author}{\bibinfo{person}{Kashyap Chitta}, \bibinfo{person}{Aditya
  Prakash}, \bibinfo{person}{Bernhard Jaeger}, \bibinfo{person}{Zehao Yu},
  \bibinfo{person}{Katrin Renz}, {and} \bibinfo{person}{Andreas Geiger}.}
  \bibinfo{year}{2022}\natexlab{}.
\newblock \showarticletitle{TransFuser: Imitation with Transformer-Based Sensor
  Fusion for Autonomous Driving}.
\newblock \bibinfo{journal}{\emph{arXiv preprint arXiv:2205.15997}}
  (\bibinfo{year}{2022}).
\newblock


\bibitem[Chu et~al\mbox{.}(2023)]%
        {chu2023review}
\bibfield{author}{\bibinfo{person}{Hongqing Chu}, \bibinfo{person}{Hejian
  Zhuang}, \bibinfo{person}{Wenshuo Wang}, \bibinfo{person}{Xiaoxiang Na},
  \bibinfo{person}{Lulu Guo}, \bibinfo{person}{Jia Zhang},
  \bibinfo{person}{Bingzhao Gao}, {and} \bibinfo{person}{Hong Chen}.}
  \bibinfo{year}{2023}\natexlab{}.
\newblock \showarticletitle{A Review of Driving Style Recognition Methods From
  Short-Term and Long-Term Perspectives}.
\newblock \bibinfo{journal}{\emph{IEEE Transactions on Intelligent Vehicles}}
  \bibinfo{volume}{8}, \bibinfo{number}{11} (\bibinfo{year}{2023}),
  \bibinfo{pages}{4599--4612}.
\newblock


\bibitem[{comma.ai}(2024)]%
        {openpilot}
\bibfield{author}{\bibinfo{person}{{comma.ai}}.}
  \bibinfo{year}{2024}\natexlab{}.
\newblock \bibinfo{title}{openpilot: an open source advanced driver assistance
  system}.
\newblock \bibinfo{howpublished}{\url{https://github.com/commaai/openpilot}}.
\newblock


\bibitem[{comma.ai}(2026)]%
        {comma_device}
\bibfield{author}{\bibinfo{person}{{comma.ai}}.}
  \bibinfo{year}{2026}\natexlab{}.
\newblock \bibinfo{title}{comma: In-Vehicle Devices for Running openpilot}.
\newblock \bibinfo{howpublished}{\url{https://comma.ai/}}.
\newblock
\shownote{Accessed: 2026-07-25}.
\newblock


\bibitem[Deng et~al\mbox{.}(2018)]%
        {arcface}
\bibfield{author}{\bibinfo{person}{Jiankang Deng}, \bibinfo{person}{Jia Guo},
  \bibinfo{person}{Jing Yang}, \bibinfo{person}{Niannan Xue},
  \bibinfo{person}{Irene Kotsia}, {and} \bibinfo{person}{Stefanos Zafeiriou}.}
  \bibinfo{year}{2018}\natexlab{}.
\newblock \showarticletitle{ArcFace: Additive Angular Margin Loss for Deep Face
  Recognition}.
\newblock \bibinfo{journal}{\emph{arXiv preprint arXiv:1801.07698}}
  (\bibinfo{year}{2018}).
\newblock


\bibitem[Dong et~al\mbox{.}(2026)]%
        {person2drive}
\bibfield{author}{\bibinfo{person}{Xiaoru Dong}, \bibinfo{person}{Ruiqin Li},
  \bibinfo{person}{Xiao Han}, \bibinfo{person}{Zhenxuan Wu},
  \bibinfo{person}{Jiamin Wang}, \bibinfo{person}{Jian Chen},
  \bibinfo{person}{Qi Jiang}, \bibinfo{person}{SM Yiu}, \bibinfo{person}{Xinge
  Zhu}, {and} \bibinfo{person}{Yuexin Ma}.} \bibinfo{year}{2026}\natexlab{}.
\newblock \showarticletitle{Driving with A Thousand Faces: A Benchmark for
  Closed-Loop Personalized End-to-End Autonomous Driving}.
\newblock \bibinfo{journal}{\emph{arXiv preprint arXiv:2602.18757}}
  (\bibinfo{year}{2026}).
\newblock


\bibitem[Enev et~al\mbox{.}(2016)]%
        {enev2016automobile}
\bibfield{author}{\bibinfo{person}{Miro Enev}, \bibinfo{person}{Alex Takakuwa},
  \bibinfo{person}{Karl Koscher}, {and} \bibinfo{person}{Tadayoshi Kohno}.}
  \bibinfo{year}{2016}\natexlab{}.
\newblock \showarticletitle{Automobile driver fingerprinting}.
\newblock \bibinfo{journal}{\emph{Proceedings on Privacy Enhancing
  Technologies}} (\bibinfo{year}{2016}).
\newblock


\bibitem[Ganin et~al\mbox{.}(2015)]%
        {dann}
\bibfield{author}{\bibinfo{person}{Yaroslav Ganin}, \bibinfo{person}{Evgeniya
  Ustinova}, \bibinfo{person}{Hana Ajakan}, \bibinfo{person}{Pascal Germain},
  \bibinfo{person}{Hugo Larochelle}, {and} \bibinfo{person}{François
  Laviolette}.} \bibinfo{year}{2015}\natexlab{}.
\newblock \showarticletitle{Domain-Adversarial Training of Neural Networks}.
\newblock \bibinfo{journal}{\emph{arXiv preprint arXiv:1505.07818}}
  (\bibinfo{year}{2015}).
\newblock


\bibitem[Gebru et~al\mbox{.}(2021)]%
        {gebru2021datasheets}
\bibfield{author}{\bibinfo{person}{Timnit Gebru}, \bibinfo{person}{Jamie
  Morgenstern}, \bibinfo{person}{Briana Vecchione},
  \bibinfo{person}{Jennifer~Wortman Vaughan}, \bibinfo{person}{Hanna Wallach},
  \bibinfo{person}{Hal~Daum{\'e} Iii}, {and} \bibinfo{person}{Kate Crawford}.}
  \bibinfo{year}{2021}\natexlab{}.
\newblock \showarticletitle{Datasheets for datasets}.
\newblock \bibinfo{journal}{\emph{Commun. ACM}} \bibinfo{volume}{64},
  \bibinfo{number}{12} (\bibinfo{year}{2021}), \bibinfo{pages}{86--92}.
\newblock


\bibitem[Geirhos et~al\mbox{.}(2020)]%
        {geirhos2020shortcut}
\bibfield{author}{\bibinfo{person}{Robert Geirhos},
  \bibinfo{person}{J{\"o}rn-Henrik Jacobsen}, \bibinfo{person}{Claudio
  Michaelis}, \bibinfo{person}{Richard Zemel}, \bibinfo{person}{Wieland
  Brendel}, \bibinfo{person}{Matthias Bethge}, {and} \bibinfo{person}{Felix~A.
  Wichmann}.} \bibinfo{year}{2020}\natexlab{}.
\newblock \showarticletitle{Shortcut Learning in Deep Neural Networks}.
\newblock \bibinfo{journal}{\emph{Nature Machine Intelligence}}
  \bibinfo{volume}{2}, \bibinfo{number}{11} (\bibinfo{year}{2020}),
  \bibinfo{pages}{665--673}.
\newblock


\bibitem[Goswami et~al\mbox{.}(2024)]%
        {moment}
\bibfield{author}{\bibinfo{person}{Mononito Goswami}, \bibinfo{person}{Konrad
  Szafer}, \bibinfo{person}{Arjun Choudhry}, \bibinfo{person}{Yifu Cai},
  \bibinfo{person}{Shuo Li}, {and} \bibinfo{person}{Artur Dubrawski}.}
  \bibinfo{year}{2024}\natexlab{}.
\newblock \showarticletitle{MOMENT: A Family of Open Time-series Foundation
  Models}.
\newblock \bibinfo{journal}{\emph{arXiv preprint arXiv:2402.03885}}
  (\bibinfo{year}{2024}).
\newblock


\bibitem[Hao et~al\mbox{.}(2025)]%
        {styledrive}
\bibfield{author}{\bibinfo{person}{Ruiyang Hao}, \bibinfo{person}{Bowen Jing},
  \bibinfo{person}{Haibao Yu}, {and} \bibinfo{person}{Zaiqing Nie}.}
  \bibinfo{year}{2025}\natexlab{}.
\newblock \showarticletitle{StyleDrive: Towards Driving-Style Aware
  Benchmarking of End-To-End Autonomous Driving}.
\newblock \bibinfo{journal}{\emph{arXiv preprint arXiv:2506.23982}}
  (\bibinfo{year}{2025}).
\newblock


\bibitem[Heigold et~al\mbox{.}(2016)]%
        {heigold2016endtoend}
\bibfield{author}{\bibinfo{person}{Georg Heigold}, \bibinfo{person}{Ignacio
  Moreno}, \bibinfo{person}{Samy Bengio}, {and} \bibinfo{person}{Noam
  Shazeer}.} \bibinfo{year}{2016}\natexlab{}.
\newblock \showarticletitle{End-to-End Text-Dependent Speaker Verification}. In
  \bibinfo{booktitle}{\emph{Proceedings of the IEEE International Conference on
  Acoustics, Speech and Signal Processing (ICASSP)}}.
  \bibinfo{pages}{5115--5119}.
\newblock


\bibitem[Huh et~al\mbox{.}(2024)]%
        {huh2024voxsrc}
\bibfield{author}{\bibinfo{person}{Jaesung Huh}, \bibinfo{person}{Joon~Son
  Chung}, \bibinfo{person}{Arsha Nagrani}, \bibinfo{person}{Andrew Brown},
  \bibinfo{person}{Jee weon Jung}, \bibinfo{person}{Daniel Garcia-Romero},
  {and} \bibinfo{person}{Andrew Zisserman}.} \bibinfo{year}{2024}\natexlab{}.
\newblock \showarticletitle{The VoxCeleb Speaker Recognition Challenge: A
  Retrospective}.
\newblock \bibinfo{journal}{\emph{IEEE/ACM Transactions on Audio, Speech, and
  Language Processing}} (\bibinfo{year}{2024}).
\newblock
\href{https://doi.org/10.1109/TASLP.2024.3444456}{doi:\nolinkurl{10.1109/TASLP.2024.3444456}}


\bibitem[Itkonen et~al\mbox{.}(2020)]%
        {itkonen2020characterisation}
\bibfield{author}{\bibinfo{person}{Teemu~H. Itkonen}, \bibinfo{person}{Esko
  Lehtonen}, {and} \bibinfo{person}{Selpi}.} \bibinfo{year}{2020}\natexlab{}.
\newblock \showarticletitle{Characterisation of Motorway Driving Style Using
  Naturalistic Driving Data}.
\newblock \bibinfo{journal}{\emph{Transportation Research Part F: Traffic
  Psychology and Behaviour}}  \bibinfo{volume}{69} (\bibinfo{year}{2020}),
  \bibinfo{pages}{72--79}.
\newblock
\href{https://doi.org/10.1016/j.trf.2020.01.003}{doi:\nolinkurl{10.1016/j.trf.2020.01.003}}


\bibitem[Khosla et~al\mbox{.}(2020)]%
        {supcon}
\bibfield{author}{\bibinfo{person}{Prannay Khosla}, \bibinfo{person}{Piotr
  Teterwak}, \bibinfo{person}{Chen Wang}, \bibinfo{person}{Aaron Sarna},
  \bibinfo{person}{Yonglong Tian}, {and} \bibinfo{person}{Phillip Isola}.}
  \bibinfo{year}{2020}\natexlab{}.
\newblock \showarticletitle{Supervised Contrastive Learning}.
\newblock \bibinfo{journal}{\emph{arXiv preprint arXiv:2004.11362}}
  (\bibinfo{year}{2020}).
\newblock


\bibitem[Krajewski et~al\mbox{.}(2018)]%
        {highd}
\bibfield{author}{\bibinfo{person}{Robert Krajewski}, \bibinfo{person}{Julian
  Bock}, \bibinfo{person}{Laurent Kloeker}, {and} \bibinfo{person}{Lutz
  Eckstein}.} \bibinfo{year}{2018}\natexlab{}.
\newblock \showarticletitle{The highD Dataset: A Drone Dataset of Naturalistic
  Vehicle Trajectories on German Highways for Validation of Highly Automated
  Driving Systems}.
\newblock \bibinfo{journal}{\emph{arXiv preprint arXiv:1810.05642}}
  (\bibinfo{year}{2018}).
\newblock


\bibitem[Liao et~al\mbox{.}(2025a)]%
        {diffusiondrive}
\bibfield{author}{\bibinfo{person}{Bencheng Liao}, \bibinfo{person}{Shaoyu
  Chen}, \bibinfo{person}{Haoran Yin}, \bibinfo{person}{Bo Jiang},
  \bibinfo{person}{Cheng Wang}, \bibinfo{person}{Sixu Yan},
  \bibinfo{person}{Xinbang Zhang}, \bibinfo{person}{Xiangyu Li},
  \bibinfo{person}{Ying Zhang}, \bibinfo{person}{Qian Zhang}, {and}
  \bibinfo{person}{Xinggang Wang}.} \bibinfo{year}{2025}\natexlab{a}.
\newblock \showarticletitle{DiffusionDrive: Truncated Diffusion Model for
  End-to-End Autonomous Driving}. In \bibinfo{booktitle}{\emph{Proceedings of
  the IEEE/CVF Conference on Computer Vision and Pattern Recognition}}.
  \bibinfo{pages}{12037--12047}.
\newblock
\href{https://doi.org/10.1109/CVPR52734.2025.01124}{doi:\nolinkurl{10.1109/CVPR52734.2025.01124}}


\bibitem[Liao et~al\mbox{.}(2023)]%
        {liao2023ddt}
\bibfield{author}{\bibinfo{person}{Xishun Liao}, \bibinfo{person}{Xuanpeng
  Zhao}, \bibinfo{person}{Ziran Wang}, \bibinfo{person}{Zhouqiao Zhao},
  \bibinfo{person}{Kyungtae Han}, \bibinfo{person}{Rohit Gupta},
  \bibinfo{person}{Matthew~J. Barth}, {and} \bibinfo{person}{Guoyuan Wu}.}
  \bibinfo{year}{2023}\natexlab{}.
\newblock \showarticletitle{Driver Digital Twin for Online Prediction of
  Personalized Lane-Change Behavior}.
\newblock \bibinfo{journal}{\emph{IEEE Internet of Things Journal}}
  \bibinfo{volume}{10}, \bibinfo{number}{15} (\bibinfo{year}{2023}),
  \bibinfo{pages}{13235--13246}.
\newblock
\href{https://doi.org/10.1109/JIOT.2023.3262484}{doi:\nolinkurl{10.1109/JIOT.2023.3262484}}


\bibitem[Liao et~al\mbox{.}(2025b)]%
        {liao2025personalization}
\bibfield{author}{\bibinfo{person}{Xishun Liao}, \bibinfo{person}{Zhouqiao
  Zhao}, \bibinfo{person}{Matthew~J. Barth}, \bibinfo{person}{Amr Abdelraouf},
  \bibinfo{person}{Rohit Gupta}, \bibinfo{person}{Kyungtae Han},
  \bibinfo{person}{Jiaqi Ma}, {and} \bibinfo{person}{Guoyuan Wu}.}
  \bibinfo{year}{2025}\natexlab{b}.
\newblock \showarticletitle{A Review of Personalization in Driving Behavior:
  Dataset, Modeling, and Validation}.
\newblock \bibinfo{journal}{\emph{IEEE Transactions on Intelligent Vehicles}}
  \bibinfo{volume}{10}, \bibinfo{number}{2} (\bibinfo{year}{2025}),
  \bibinfo{pages}{1241--1262}.
\newblock
\href{https://doi.org/10.1109/TIV.2024.3425647}{doi:\nolinkurl{10.1109/TIV.2024.3425647}}


\bibitem[Liu et~al\mbox{.}(2023)]%
        {itransformer}
\bibfield{author}{\bibinfo{person}{Yong Liu}, \bibinfo{person}{Tengge Hu},
  \bibinfo{person}{Haoran Zhang}, \bibinfo{person}{Haixu Wu},
  \bibinfo{person}{Shiyu Wang}, {and} \bibinfo{person}{Lintao Ma}.}
  \bibinfo{year}{2023}\natexlab{}.
\newblock \showarticletitle{iTransformer: Inverted Transformers Are Effective
  for Time Series Forecasting}.
\newblock \bibinfo{journal}{\emph{arXiv preprint arXiv:2310.06625}}
  (\bibinfo{year}{2023}).
\newblock


\bibitem[Lyu et~al\mbox{.}(2022)]%
        {lyu2022naturalistic}
\bibfield{author}{\bibinfo{person}{Nengchao Lyu}, \bibinfo{person}{Yugang
  Wang}, \bibinfo{person}{Chaozhong Wu}, \bibinfo{person}{Lingfeng Peng}, {and}
  \bibinfo{person}{Alieu~Freddie Thomas}.} \bibinfo{year}{2022}\natexlab{}.
\newblock \showarticletitle{Using Naturalistic Driving Data to Identify Driving
  Style Based on Longitudinal Driving Operation Conditions}.
\newblock \bibinfo{journal}{\emph{Journal of Intelligent and Connected
  Vehicles}} \bibinfo{volume}{5}, \bibinfo{number}{1} (\bibinfo{year}{2022}),
  \bibinfo{pages}{17--35}.
\newblock
\href{https://doi.org/10.1108/JICV-07-2021-0008}{doi:\nolinkurl{10.1108/JICV-07-2021-0008}}


\bibitem[Nie et~al\mbox{.}(2022)]%
        {patchtst}
\bibfield{author}{\bibinfo{person}{Yuqi Nie}, \bibinfo{person}{Nam~H. Nguyen},
  \bibinfo{person}{Phanwadee Sinthong}, {and} \bibinfo{person}{Jayant
  Kalagnanam}.} \bibinfo{year}{2022}\natexlab{}.
\newblock \showarticletitle{A Time Series is Worth 64 Words: Long-term
  Forecasting with Transformers}.
\newblock \bibinfo{journal}{\emph{arXiv preprint arXiv:2211.14730}}
  (\bibinfo{year}{2022}).
\newblock


\bibitem[Oquab et~al\mbox{.}(2023)]%
        {dinov2}
\bibfield{author}{\bibinfo{person}{Maxime Oquab}, \bibinfo{person}{Timothée
  Darcet}, \bibinfo{person}{Théo Moutakanni}, \bibinfo{person}{Huy Vo},
  \bibinfo{person}{Marc Szafraniec}, {and} \bibinfo{person}{Vasil Khalidov}.}
  \bibinfo{year}{2023}\natexlab{}.
\newblock \showarticletitle{DINOv2: Learning Robust Visual Features without
  Supervision}.
\newblock \bibinfo{journal}{\emph{arXiv preprint arXiv:2304.07193}}
  (\bibinfo{year}{2023}).
\newblock


\bibitem[Perez et~al\mbox{.}(2017)]%
        {film}
\bibfield{author}{\bibinfo{person}{Ethan Perez}, \bibinfo{person}{Florian
  Strub}, \bibinfo{person}{Harm de Vries}, \bibinfo{person}{Vincent Dumoulin},
  {and} \bibinfo{person}{Aaron Courville}.} \bibinfo{year}{2017}\natexlab{}.
\newblock \showarticletitle{FiLM: Visual Reasoning with a General Conditioning
  Layer}.
\newblock \bibinfo{journal}{\emph{arXiv preprint arXiv:1709.07871}}
  (\bibinfo{year}{2017}).
\newblock


\bibitem[Qiu et~al\mbox{.}(2026)]%
        {plans}
\bibfield{author}{\bibinfo{person}{Xiaoyun Qiu}, \bibinfo{person}{Jingtao He},
  \bibinfo{person}{Yijie Chen}, \bibinfo{person}{Yusong Huang},
  \bibinfo{person}{Haotian Wang}, \bibinfo{person}{Yixuan Wang}, {and}
  \bibinfo{person}{Xinhu Zheng}.} \bibinfo{year}{2026}\natexlab{}.
\newblock \showarticletitle{{PLAN-S}: Bridging Planning with Latent Style
  Dynamics for Autonomous Driving World Models}.
\newblock \bibinfo{journal}{\emph{arXiv preprint arXiv:2606.06014}}
  (\bibinfo{year}{2026}).
\newblock


\bibitem[Ramanishka et~al\mbox{.}(2018)]%
        {hdd}
\bibfield{author}{\bibinfo{person}{Vasili Ramanishka}, \bibinfo{person}{Yi-Ting
  Chen}, \bibinfo{person}{Teruhisa Misu}, {and} \bibinfo{person}{Kate Saenko}.}
  \bibinfo{year}{2018}\natexlab{}.
\newblock \showarticletitle{Toward Driving Scene Understanding: A Dataset for
  Learning Driver Behavior and Causal Reasoning}. In
  \bibinfo{booktitle}{\emph{Proceedings of the IEEE Conference on Computer
  Vision and Pattern Recognition}}. \bibinfo{pages}{7699--7707}.
\newblock
\href{https://doi.org/10.1109/CVPR.2018.00803}{doi:\nolinkurl{10.1109/CVPR.2018.00803}}


\bibitem[Remeli et~al\mbox{.}(2019)]%
        {remeli2019automatic}
\bibfield{author}{\bibinfo{person}{Mina Remeli}, \bibinfo{person}{Szilvia
  Lestyan}, \bibinfo{person}{Gergely Acs}, {and} \bibinfo{person}{Gergely
  Biczok}.} \bibinfo{year}{2019}\natexlab{}.
\newblock \showarticletitle{Automatic Driver Identification from In-Vehicle
  Network Logs}.
\newblock \bibinfo{journal}{\emph{arXiv preprint arXiv:1911.09508}}
  (\bibinfo{year}{2019}).
\newblock


\bibitem[Sagberg et~al\mbox{.}(2015)]%
        {sagberg2015review}
\bibfield{author}{\bibinfo{person}{Fridulv Sagberg}, \bibinfo{person}{Selpi},
  \bibinfo{person}{Giulio F.~Bianchi Piccinini}, {and} \bibinfo{person}{Johan
  Engstr{\"o}m}.} \bibinfo{year}{2015}\natexlab{}.
\newblock \showarticletitle{A Review of Research on Driving Styles and Road
  Safety}.
\newblock \bibinfo{journal}{\emph{Human Factors}} \bibinfo{volume}{57},
  \bibinfo{number}{7} (\bibinfo{year}{2015}), \bibinfo{pages}{1248--1275}.
\newblock
\href{https://doi.org/10.1177/0018720815591313}{doi:\nolinkurl{10.1177/0018720815591313}}


\bibitem[Schafer et~al\mbox{.}(2018)]%
        {comma2k19}
\bibfield{author}{\bibinfo{person}{Harald Schafer}, \bibinfo{person}{Eder
  Santana}, \bibinfo{person}{Andrew Haden}, {and} \bibinfo{person}{Riccardo
  Biasini}.} \bibinfo{year}{2018}\natexlab{}.
\newblock \showarticletitle{A Commute in Data: The comma2k19 Dataset}.
\newblock \bibinfo{journal}{\emph{arXiv preprint arXiv:1812.05752}}
  (\bibinfo{year}{2018}).
\newblock


\bibitem[Siméoni et~al\mbox{.}(2025)]%
        {dinov3}
\bibfield{author}{\bibinfo{person}{Oriane Siméoni}, \bibinfo{person}{Huy~V.
  Vo}, \bibinfo{person}{Maximilian Seitzer}, \bibinfo{person}{Federico
  Baldassarre}, \bibinfo{person}{Maxime Oquab}, {and} \bibinfo{person}{Cijo
  Jose}.} \bibinfo{year}{2025}\natexlab{}.
\newblock \showarticletitle{DINOv3}.
\newblock \bibinfo{journal}{\emph{arXiv preprint arXiv:2508.10104}}
  (\bibinfo{year}{2025}).
\newblock


\bibitem[Snell et~al\mbox{.}(2017)]%
        {protonet}
\bibfield{author}{\bibinfo{person}{Jake Snell}, \bibinfo{person}{Kevin
  Swersky}, {and} \bibinfo{person}{Richard~S. Zemel}.}
  \bibinfo{year}{2017}\natexlab{}.
\newblock \showarticletitle{Prototypical Networks for Few-shot Learning}.
\newblock \bibinfo{journal}{\emph{arXiv preprint arXiv:1703.05175}}
  (\bibinfo{year}{2017}).
\newblock


\bibitem[Sun et~al\mbox{.}(2020)]%
        {sun2020scalability}
\bibfield{author}{\bibinfo{person}{Pei Sun}, \bibinfo{person}{Henrik
  Kretzschmar}, \bibinfo{person}{Xerxes Dotiwalla}, \bibinfo{person}{Aurelien
  Chouard}, \bibinfo{person}{Vijaysai Patnaik}, \bibinfo{person}{Paul Tsui},
  \bibinfo{person}{James Guo}, \bibinfo{person}{Yin Zhou},
  \bibinfo{person}{Yuning Chai}, \bibinfo{person}{Benjamin Caine},
  {et~al\mbox{.}}} \bibinfo{year}{2020}\natexlab{}.
\newblock \showarticletitle{Scalability in perception for autonomous driving:
  Waymo open dataset}. In \bibinfo{booktitle}{\emph{Proceedings of the IEEE/CVF
  conference on computer vision and pattern recognition}}.
  \bibinfo{pages}{2446--2454}.
\newblock


\bibitem[Tschannen et~al\mbox{.}(2025)]%
        {siglip2}
\bibfield{author}{\bibinfo{person}{Michael Tschannen}, \bibinfo{person}{Alexey
  Gritsenko}, \bibinfo{person}{Xiao Wang}, \bibinfo{person}{Muhammad~Ferjad
  Naeem}, \bibinfo{person}{Ibrahim Alabdulmohsin}, {and}
  \bibinfo{person}{Nikhil Parthasarathy}.} \bibinfo{year}{2025}\natexlab{}.
\newblock \showarticletitle{SigLIP 2: Multilingual Vision-Language Encoders
  with Improved Semantic Understanding, Localization, and Dense Features}.
\newblock \bibinfo{journal}{\emph{arXiv preprint arXiv:2502.14786}}
  (\bibinfo{year}{2025}).
\newblock


\bibitem[Tselentis and Papadimitriou(2023)]%
        {tselentis2023driver}
\bibfield{author}{\bibinfo{person}{Dimitrios~I. Tselentis} {and}
  \bibinfo{person}{Eleonora Papadimitriou}.} \bibinfo{year}{2023}\natexlab{}.
\newblock \showarticletitle{Driver Profile and Driving Pattern Recognition for
  Road Safety Assessment: Main Challenges and Future Directions}.
\newblock \bibinfo{journal}{\emph{IEEE Open Journal of Intelligent
  Transportation Systems}}  \bibinfo{volume}{4} (\bibinfo{year}{2023}),
  \bibinfo{pages}{83--100}.
\newblock
\href{https://doi.org/10.1109/OJITS.2023.3237177}{doi:\nolinkurl{10.1109/OJITS.2023.3237177}}


\bibitem[{Virginia Tech Transportation Institute}(2025)]%
        {shrp2nds}
\bibfield{author}{\bibinfo{person}{{Virginia Tech Transportation Institute}}.}
  \bibinfo{year}{2025}\natexlab{}.
\newblock \bibinfo{title}{{SHRP2 Naturalistic Driving Study Data Access}}.
\newblock \bibinfo{howpublished}{Online database}.
\newblock
\shownote{Accessed 2026-07-21}.
\newblock
\urldef\tempurl%
\url{https://insight.shrp2nds.us/}
\showURL{%
\tempurl}


\bibitem[Wei et~al\mbox{.}(2025)]%
        {pdb}
\bibfield{author}{\bibinfo{person}{Chuheng Wei}, \bibinfo{person}{Ziye Qin},
  \bibinfo{person}{Siyan Li}, {and} \bibinfo{person}{Ziyan Zhang}.}
  \bibinfo{year}{2025}\natexlab{}.
\newblock \showarticletitle{PDB: Not All Drivers Are the Same -- A Personalized
  Dataset for Understanding Driving Behavior}.
\newblock \bibinfo{journal}{\emph{arXiv preprint arXiv:2503.06477}}
  (\bibinfo{year}{2025}).
\newblock


\bibitem[Wen et~al\mbox{.}(2022)]%
        {wen2022carfollowing}
\bibfield{author}{\bibinfo{person}{Xiao Wen}, \bibinfo{person}{Zhiyong Cui},
  {and} \bibinfo{person}{Sisi Jian}.} \bibinfo{year}{2022}\natexlab{}.
\newblock \showarticletitle{Characterizing Car-Following Behaviors of Human
  Drivers When Following Automated Vehicles Using the Real-World Dataset}.
\newblock \bibinfo{journal}{\emph{Accident Analysis \& Prevention}}
  \bibinfo{volume}{172} (\bibinfo{year}{2022}), \bibinfo{pages}{106689}.
\newblock
\href{https://doi.org/10.1016/j.aap.2022.106689}{doi:\nolinkurl{10.1016/j.aap.2022.106689}}


\bibitem[Wu(2025)]%
        {pdbeval}
\bibfield{author}{\bibinfo{person}{Junda Wu}.} \bibinfo{year}{2025}\natexlab{}.
\newblock \showarticletitle{PDB-Eval: An Evaluation of Large Multimodal Models
  for Description and Explanation of Personalized Driving Behavior}.
\newblock \bibinfo{journal}{\emph{arXiv preprint arXiv:2507.18447}}
  (\bibinfo{year}{2025}).
\newblock


\bibitem[Yang et~al\mbox{.}(2020)]%
        {yang2020driver2vec}
\bibfield{author}{\bibinfo{person}{Jingbo Yang}, \bibinfo{person}{Ruge Zhao},
  \bibinfo{person}{Meixian Zhu}, \bibinfo{person}{David Hallac},
  \bibinfo{person}{Jaka Sodnik}, {and} \bibinfo{person}{Jure Leskovec}.}
  \bibinfo{year}{2020}\natexlab{}.
\newblock \showarticletitle{{Driver2vec}: Driver Identification from Automotive
  Data}. In \bibinfo{booktitle}{\emph{Proceedings of the 6th Workshop on Mining
  and Learning from Time Series ({MiLeTS} 2020), Co-located with {KDD} 2020}}.
  \bibinfo{address}{San Diego, California, USA}.
\newblock


\bibitem[Yang et~al\mbox{.}(2021)]%
        {driver2vec}
\bibfield{author}{\bibinfo{person}{Jingbo Yang}, \bibinfo{person}{Ruge Zhao},
  \bibinfo{person}{Meixian Zhu}, \bibinfo{person}{David Hallac},
  \bibinfo{person}{Jaka Sodnik}, {and} \bibinfo{person}{Jure Leskovec}.}
  \bibinfo{year}{2021}\natexlab{}.
\newblock \showarticletitle{Driver2vec: Driver Identification from Automotive
  Data}.
\newblock \bibinfo{journal}{\emph{arXiv preprint arXiv:2102.05234}}
  (\bibinfo{year}{2021}).
\newblock


\bibitem[Yu et~al\mbox{.}(2018)]%
        {bdd100k}
\bibfield{author}{\bibinfo{person}{Fisher Yu}, \bibinfo{person}{Haofeng Chen},
  \bibinfo{person}{Xin Wang}, \bibinfo{person}{Wenqi Xian},
  \bibinfo{person}{Yingying Chen}, {and} \bibinfo{person}{Fangchen Liu}.}
  \bibinfo{year}{2018}\natexlab{}.
\newblock \showarticletitle{BDD100K: A Diverse Driving Dataset for
  Heterogeneous Multitask Learning}.
\newblock \bibinfo{journal}{\emph{arXiv preprint arXiv:1805.04687}}
  (\bibinfo{year}{2018}).
\newblock


\bibitem[Yue et~al\mbox{.}(2021)]%
        {ts2vec}
\bibfield{author}{\bibinfo{person}{Zhihan Yue}, \bibinfo{person}{Yujing Wang},
  \bibinfo{person}{Juanyong Duan}, \bibinfo{person}{Tianmeng Yang},
  \bibinfo{person}{Congrui Huang}, {and} \bibinfo{person}{Yunhai Tong}.}
  \bibinfo{year}{2021}\natexlab{}.
\newblock \showarticletitle{TS2Vec: Towards Universal Representation of Time
  Series}.
\newblock \bibinfo{journal}{\emph{arXiv preprint arXiv:2106.10466}}
  (\bibinfo{year}{2021}).
\newblock


\end{thebibliography}

\clearpage
\appendix
\section*{Appendix}

\section{Decoding and the Log-Tier Confound}
\label{app:decode}

The fleet uploads two log tiers: a compact tier present on nearly all drives and a full-rate tier present on a subset.
Which tier a drive carries correlates with device firmware and vehicle platform.
In an early pilot we decoded whichever tier was available per drive.
A linear probe then predicted the log tier from a window embedding at AUROC 0.607, and vehicle-model leakage rose with it.
After re-decoding every drive from the compact tier at a uniform 10\,Hz, the same probe returned AUROC 0.500.
All released data uses the uniform decode.
Signal-fidelity checks (per-channel standard-deviation difference below 0.2\% against the full-rate tier; realized curvature consistent with yaw-rate-derived curvature) are included in the release.

\section{Scenario Taxonomy and Behavioral Primitives}
\label{app:taxonomy}

The main text states that scenarios and primitives are rule-derived; this section gives the exact rules.

\textbf{Scenario taxonomy.}
Each 60\,s window receives one scenario label by deterministic priority rules evaluated in order:
stop-and-go if the stopped fraction exceeds 0.10;
otherwise curve if the 95th-percentile realized curvature exceeds 0.01\,m$^{-1}$;
otherwise car-following if a lead vehicle is present for more than half the window and median THW is below 3.5\,s;
otherwise high-speed if mean speed exceeds 25\,m/s;
otherwise urban if mean speed is below 15\,m/s;
otherwise free driving.
The priority order resolves overlaps (e.g., a stop-and-go window on a curved road is stop-and-go).

\textbf{Behavioral primitives.}
Each primitive is a scenario-conditioned percentile label on one window statistic (Table~\ref{tab:primitives}): a window is labeled high when its indicator is at or above the 80th percentile \emph{within its own scenario type}, low at or below the 20th, and neutral otherwise.
Steering comes in two versions---raw steering-rate (vehicle-dependent through the steering ratio) and curvature-rate (vehicle-normalized)---and the pair feeds the vehicle-leakage diagnostic in Section~\ref{sec:results}.
Curve-entry deceleration is evaluated within the curve scenario only.

\begin{table}[htbp]
\centering
\caption{Behavioral-primitive definitions. Labels are Q80/Q20 percentile cuts of the indicator within each scenario type.}
\label{tab:primitives}
\footnotesize
\setlength{\tabcolsep}{3.8pt}
\renewcommand{\arraystretch}{1.12}
\begin{tabularx}{\columnwidth}{@{}
  >{\raggedright\arraybackslash}p{0.31\columnwidth}
  >{\raggedright\arraybackslash}p{0.34\columnwidth}
  >{\raggedright\arraybackslash}X@{}}
\toprule
\rowcolor{DriveTeal}
\color{white}\textbf{Primitive} &
\color{white}\textbf{Indicator} &
\color{white}\textbf{High label} \\
\midrule
close following & median THW & low tail \\
large headway & median THW & high tail \\
hard braking & 5th-pct. acceleration & low (more negative) \\
high jerk & jerk RMS & high tail \\
sharp steering (input) & steering-rate RMS & high tail \\
sharp steering (path) & curvature-rate RMS & high tail \\
lane correction & lane-offset SDLP & high tail \\
\rowcolor{DriveMint!38}
curve-entry decel. & 5th-pct. accel. (curve only) & low (more negative) \\
\bottomrule
\end{tabularx}
\end{table}

\section{Annotation Audit and Threshold Sensitivity}
\label{app:annotation}

Moving the percentile thresholds from Q80/Q20 to Q75/Q85 changes label assignments at the boundary but leaves the downstream conclusions unchanged: re-identification metrics move by less than one seed standard deviation, and the sign of every stratified personalization gain is preserved.

A human audit covered 240 windows (8 primitives $\times$ \{high, low\} $\times$ 15, stratified and shuffled).
For each window the auditor watched the full 60\,s forward video next to the aligned signal traces and judged whether the label matches the observed driving.
Overall agreement is \textbf{93.0\%} (excluding 12 unsure votes); Table~\ref{tab:audit} reports per-primitive results.
Agreement is lowest for large-headway (84\%), the most subjective judgment, and highest for curve-entry deceleration (100\%).
The audit interface and raw answers ship with the release.

\begin{table}[htbp]
\centering
\caption{Human audit of annotation labels (30 windows per primitive). Agreement excludes unsure votes.}
\label{tab:audit}
\footnotesize
\setlength{\tabcolsep}{2.8pt}
\renewcommand{\arraystretch}{1.12}
\begin{tabularx}{\columnwidth}{@{}
  >{\raggedright\arraybackslash}X
  >{\centering\arraybackslash}p{0.10\columnwidth}
  >{\centering\arraybackslash}p{0.10\columnwidth}
  >{\centering\arraybackslash}p{0.10\columnwidth}
  >{\centering\arraybackslash}p{0.17\columnwidth}@{}}
\toprule
\rowcolor{DriveTeal}
\color{white}\textbf{Primitive} &
\color{white}\textbf{Correct} &
\color{white}\textbf{Wrong} &
\color{white}\textbf{Unsure} &
\color{white}\textbf{Agreement} \\
\midrule
close following & 28 & 2 & 0 & 93.3\% \\
large headway & 21 & 4 & 5 & 84.0\% \\
hard braking & 27 & 1 & 2 & 96.4\% \\
high jerk & 29 & 1 & 0 & 96.7\% \\
sharp steering (input) & 27 & 2 & 1 & 93.1\% \\
sharp steering (path) & 25 & 3 & 2 & 89.3\% \\
lane correction & 26 & 3 & 1 & 89.7\% \\
curve-entry deceleration & 29 & 0 & 1 & 100.0\% \\
\midrule
\rowcolor{DriveMint!38}
\textbf{Overall} & \textbf{212} & \textbf{16} & \textbf{12} & \textbf{93.0\%} \\
\bottomrule
\end{tabularx}
\end{table}

\section{Maneuver-Event Layer and Its Human Audit}
\label{app:maneuvers}

Beyond the behavioral primitives, the release includes a maneuver-event layer with six classes detected by physically grounded rules over the CAN signals: deceleration episodes, acceleration episodes (velocity-rise segments with a sustained-speed check), intersection turns (steering angle and steering rate jointly), road-geometry curve driving (realized curvature), car-following state (10-s chunks of stable following with a clean radar track, mean time headway attached to every chunk), and lane-change candidates (three loose lane-line and steering rules with a 15-mph whole-clip speed gate).
The rules were tuned iteratively against human review---for example, acceleration went through three definitions (thresholded acceleration, velocity-rise, velocity-rise excluding reversing shuffles) before the audit converged---and the final rules are released with the code.

Five annotators audited the layer by watching a forward-video clip for every sampled event: 1{,}000 random events per class for five classes, and an individual review of \emph{all} 36{,}488 watchable lane-change candidates (temporally overlapping candidates were merged into single watch units whose label propagates to every member, halving review effort).
Table~\ref{tab:maneuvers} reports the results.
The five sampled classes reach 94.8--99.6\% precision, which certifies the released automatic detections.
Lane changes are handled more conservatively: candidate precision is 61.2\%, so the release separates the 22{,}322 individually verified lane changes (313 drivers, 2{,}076 drives) from the 14{,}152 rejected candidates, which are released as labeled hard negatives for detector research.
The audit also localized failure modes by platform: lane-line tracking on several Ford models produces systematic false lateral drift, so lane-change precision on those vehicles is far below the fleet average---a per-platform signal-quality caveat that the release documents alongside the per-model channel-availability table.

\textbf{The audited layer as an evaluation set.}
The verified events immediately support evaluating multimodal models.
As a first use, we ask zero-shot VLMs to classify the maneuver from four frames of each event clip (300 verified events per class; chance $=16.7\%$).
Qwen3-VL-4B reaches 35.0\% accuracy (macro-F1 .27) and Qwen2.5-VL-3B 25.3\% (.18).
The error structure is informative: classes whose answer is visible in a single frame score high (car-following 98\% recall, cornering 66\% for Qwen3-VL), while classes defined by \emph{dynamics} collapse (deceleration 9\%, acceleration 1\%, lane change 1\%), with most such events misread as car-following.
Zero-shot VLMs see the scene but not the motion---consistent with the per-frame-encoder finding of Section~\ref{sec:results:video}, and a concrete headroom measurement that the released layer makes possible.

\begin{table}[htbp]
\centering
\caption{Human audit of the maneuver-event layer. Five classes are certified by a 1{,}000-event random sample; lane-change candidates are reviewed individually. Precision excludes unsure votes.}
\label{tab:maneuvers}
\footnotesize
\setlength{\tabcolsep}{3.5pt}
\renewcommand{\arraystretch}{1.12}
\begin{tabularx}{\columnwidth}{@{}
  >{\raggedright\arraybackslash}X
  >{\raggedleft\arraybackslash}p{0.18\columnwidth}
  >{\raggedleft\arraybackslash}p{0.18\columnwidth}
  >{\raggedleft\arraybackslash}p{0.17\columnwidth}@{}}
\toprule
\rowcolor{DriveTeal}
\color{white}\textbf{Class} &
\color{white}\textbf{Events} &
\color{white}\textbf{Audited} &
\color{white}\textbf{Precision} \\
\midrule
deceleration & 36{,}004 & 999 & 99.6\% \\
acceleration & 44{,}537 & 999 & 98.6\% \\
intersection turn & 34{,}706 & 1{,}000 & 94.8\% \\
cornering & 10{,}688 & 1{,}000 & 95.9\% \\
car-following (10-s state) & 127{,}991 & 1{,}000 & 96.8\% \\
\rowcolor{DriveGreen!12}
lane change (candidates) & 37{,}287 & 36{,}488 & 61.2\%$^{\dagger}$ \\
\bottomrule
\multicolumn{4}{@{}p{\columnwidth}@{}}{\scriptsize $^{\dagger}$Candidate precision from individual review of every watchable candidate; 22{,}322 verified lane changes are released as events.}
\end{tabularx}
\end{table}

\section{Benchmark Split Definitions}
\label{app:splits}

Table~\ref{tab:splitdefs} states the constraint each released split enforces; Figure~\ref{fig:benchmark_splits} in the main text gives the visual overview.
The three additional evaluations are not alternatives to the main split but progressively stricter controls: within-nameplate removes vehicle-model information, cross-vehicle tests whether a driver's signal survives a change of car, and condition matching fixes the driving conditions themselves.

\begin{table*}[htbp]
\centering
\caption{Constraints enforced by each released split. ``---'' means that the dimension is unconstrained and may repeat across the two sides of the split.}
\label{tab:splitdefs}
\footnotesize
\setlength{\tabcolsep}{4.5pt}
\renewcommand{\arraystretch}{1.12}
\begin{tabularx}{\textwidth}{@{}
  >{\raggedright\arraybackslash}p{0.18\textwidth}
  >{\raggedright\arraybackslash}p{0.15\textwidth}
  >{\raggedright\arraybackslash}p{0.13\textwidth}
  >{\raggedright\arraybackslash}p{0.14\textwidth}
  >{\raggedright\arraybackslash}X@{}}
\toprule
\rowcolor{DriveTeal}
\color{white}\textbf{Split} &
\color{white}\textbf{Drivers} &
\color{white}\textbf{Vehicles} &
\color{white}\textbf{Drives} &
\color{white}\textbf{Purpose} \\
\midrule
\rowcolor{DriveMint!38}
Main (212/45/45) & disjoint folds & --- & follow drivers & unseen-driver generalization \\
Few-shot hold-out (53) & absent from training & --- & support/query disjoint & enrollment curves on unseen drivers \\
Within-nameplate (24 models) & different & same model & disjoint & remove model-level shortcuts \\
Cross-vehicle & same driver & different & disjoint & style transfer across cars \\
Condition-matched (14{,}868 pairs) & same or different & same model & different & condition-matched comparison \\
Missing-channel & as main & as main & as main & robustness to signal availability \\
\bottomrule
\end{tabularx}
\end{table*}

\section{Condition-Matched Pair Construction and Scene-Attribute Validation}
\label{app:vlm}

\textbf{Pair construction.}
Windows are grouped into cells keyed by scenario type $\times$ speed bin ($<$10, 10--20, 20--30, $\geq$30\,m/s) $\times$ THW bin ($<$1.2, 1.2--2.0, $>$2.0\,s; defined for car-following windows only, ``n/a'' otherwise) $\times$ consolidated vehicle model.
Within each cell holding at least four windows from at least two drivers, positives pair the same driver across different drives (at most eight per driver per cell) and negatives pair different drivers; the two sets are balanced, and sampling uses a fixed seed.
This yields the 14{,}868 released pairs.

\textbf{Scene-attribute validation.}
Because the matching uses CAN variables only, we validate it visually with an independent annotator model.
Each matched-pair window is annotated from its mid-window keyframe with a fixed prompt requesting strict JSON over seven attributes (traffic density, lead vehicle, intersection, traffic signal, road type, time of day, weather).
The release uses Qwen3-VL-4B annotations (97.2\% parse success over 17{,}848 windows); an earlier Qwen2.5-VL-3B pass is included for comparison.
Matched pairs agree far more often than random pairs on the matched dimensions (road type $1.68\times$ random, traffic density $1.21\times$) and no more than random pairs on unmatched dimensions (weather $1.07\times$), consistent with the matching controlling what it claims to control and nothing more.
Between the two model generations, agreement is high on discrete attributes (time of day .93, traffic signal .89, intersection .88) and lower on graded ones (traffic density .55); the matched-pair conclusion is unchanged under either generation.

\section{Vehicle-Instance Sensitivity and Missing-Channel Robustness}
\label{app:vehinstance}

\textbf{What matching holds fixed.}
Condition-matched pairs hold the consolidated vehicle model fixed, not the physical vehicle: because most drivers are observed in a single car, same-driver positives and different-driver negatives can still differ in vehicle instance, calibration, and recording device.
Two diagnostics bound this effect indirectly: on within-nameplate splits the CAN embedding retains AUROC .887 (Section~\ref{sec:results:video}), and vehicle-model probes read $2.3\times$ chance from steering-angle inputs but chance from realized curvature (Section~\ref{sec:results:confound}), the signal family the encoder consumes.

\textbf{Within-driver cross-vehicle verification.}
The corpus contains 20 drivers observed with at least 20 windows on each of two or more consolidated models; 7 of them admit vehicle-controlled evaluation, with negatives drawn from other drivers on the query vehicle's model.
Enrolling on one vehicle and verifying on another yields mean AUROC .750 (driver-level bootstrap 95\% CI [.64, .86]); the same protocol within a single vehicle yields .816 ([.74, .90]).
The driver-associated signal therefore transfers across vehicle instances well above chance at a modest cost relative to same-vehicle verification, though the small population (7 drivers, 110 vehicle pairs) limits the precision of this comparison.

\textbf{Missing-channel robustness.}
Table~\ref{tab:missingch} removes one signal group at a time at evaluation only (values set to the training mean), with the frozen SupCon encoder.
Removing steering inputs or pedals costs about six AUROC points; removing the motion, curvature, lane, or radar group costs at most 1.3 points.
The embedding degrades gracefully rather than collapsing under any single missing group.

\begin{table}[htbp]
\centering
\caption{
Missing-channel robustness of the frozen re-identification embedding
(5-minute enrollment; 80 unseen drivers).
Single checkpoint with a fixed support draw; the three-seed full-input result is
$.935 \pm .005$.
}
\label{tab:missingch}
\footnotesize
\setlength{\tabcolsep}{4.5pt}
\renewcommand{\arraystretch}{1.12}

\begin{tabularx}{\linewidth}{@{}
  >{\raggedright\arraybackslash}X
  >{\centering\arraybackslash}p{0.18\linewidth}
  >{\centering\arraybackslash}p{0.18\linewidth}
@{}}
\toprule
\rowcolor{DriveTeal}
\color{white}\textbf{Signal group removed} &
\color{white}\textbf{AUROC} &
\color{white}\textbf{Top-1} \\
\midrule

\rowcolor{DriveMint!38}
None (full input) & .916 & .339 \\
Lead vehicle (radar) & .918 & .338 \\
Curvature / yaw & .908 & .310 \\
Lane lines & .908 & .300 \\
Motion (speed, acceleration) & .903 & .293 \\
Pedals & .859 & .222 \\
Steering input & .856 & .200 \\

\bottomrule
\end{tabularx}
\end{table}

\section{Distribution Metrics and Per-Driver Personalization Gains}
\label{app:distmetrics}

\textbf{Distribution distances.}
The distribution metrics of Section~\ref{sec:benchmark} compare predicted and observed future-behavior statistics (horizon-mean and 5th-percentile acceleration, mean absolute curvature) within scenario buckets, using one predictive sample per anchor from the CVAE head; distances are computed in normalized target units.
Personalization moves the predicted distributions toward the observed ones under MMD ($.430 \rightarrow .422$) and Wasserstein-1 ($.349 \rightarrow .339$) in each of the three seeds, while the histogram-KL shows no consistent change.

\textbf{Per-driver gains.}
Across the 80 evaluable unseen drivers and three seeds, the distributional (NLL) gain is positive for 55\% of drivers with a heavy-tailed per-driver distribution (mean $+2.4$ nats; driver-level bootstrap 95\% CI $[-0.0, 4.7]$), while the point-error gain is centered near zero ($+0.18\%$; CI $[-0.2, 0.6]$; positive for 51\% of drivers).
With this head, personalization primarily changes the predicted distribution for a subset of drivers rather than shifting the conditional mean.

\section{Audit of the Evaluation Procedure}
\label{app:protocol}

Our first seed re-runs varied a single seed that controlled both training randomness and the evaluation support/query drive split.
Under that procedure the mixture-density distributional gain read $+1.16$ nats on one seed, $+0.05$ on the second, and $-0.67$ on the third.
Freezing the evaluation split and re-running the same three training seeds gave $+0.07$, $+0.07$, and $+0.17$ nats.
The identical model that read $+1.16$ under a moving split reads $+0.068$ under the frozen split, so the original magnitude was a property of the split, not the model.
Per-window likelihoods are heavy-tailed across drivers and windows, which makes them especially sensitive to this error.
The conditional VAE shows the same pattern at a larger scale ($+6.9 \pm 4.3$ nats, positive in all seeds).
The benchmark definition therefore fixes all evaluation splits as constants and reserves seeds for training.

\begin{table}[htbp]
\centering
\caption{Baseline suite organized by five capabilities.}
\label{tab:baseline_coverage}

\setlength{\tabcolsep}{3.5pt}
\renewcommand{\arraystretch}{1.13}
\footnotesize

\begin{tabularx}{\linewidth}{
  >{\raggedright\arraybackslash}p{0.25\linewidth}
  >{\raggedright\arraybackslash}X
  >{\raggedright\arraybackslash}p{0.23\linewidth}
}
\toprule
\rowcolor{DriveTeal}
\color{white}\textbf{Baseline family} &
\color{white}\textbf{Models and methods} &
\color{white}\textbf{Purpose} \\
\midrule

\rowcolor{DriveMint!40}
\multicolumn{3}{l}{
  \textbf{1. Representation learning}
} \\

Time-series encoders &
PatchTST (joint/CI), iTransformer,
SupCon/ArcFace, ProtoNet;
MOMENT-1 zero-shot &
Modern encoders and a time-series foundation model \\

Label-free pretraining &
Masked-TS SSL;
JEPA-style latent-predictive SSL &
Value of unlabeled pretraining \\

\addlinespace[1pt]

\rowcolor{DriveBlue!22}
\multicolumn{3}{l}{
  \textbf{2. Shortcut robustness}
} \\

Leakage control &
ResidualStyle, DANN,
utility--leakage Pareto;
video-only probe &
Vehicle, route, and driving-condition shortcuts \\

\addlinespace[1pt]

\rowcolor{DriveGreen!18}
\multicolumn{3}{l}{
  \textbf{3. Personalization}
} \\

Driver conditioning &
FiLM few-shot ladder;
query-injection variant &
Ways of injecting driver information \\

\addlinespace[1pt]

\rowcolor{DriveGold!32}
\multicolumn{3}{l}{
  \textbf{4. Multimodal modeling}
} \\

Video scene information &
DINOv2/DINOv3/SigLIP2 (per-frame)
vs.\ V-JEPA~2 (temporal);
CLIP-style alignment;
VLM attributes &
Appearance, motion, and semantic scene features \\

\addlinespace[1pt]

\rowcolor{DriveTeal!16}
\multicolumn{3}{l}{
  \textbf{5. Distributional prediction}
} \\

Probabilistic heads &
MDN, CVAE &
Uncertainty and multiple behavior modes \\

\bottomrule
\end{tabularx}
\end{table}

\section{Baseline Index}
\label{app:index}

Table~\ref{tab:baseline_index} lists every evaluated baseline and where its numbers are reported, making the baseline count auditable at a glance.

\begin{table}[htbp]
\centering
\caption{Every evaluated baseline and where its results appear. Related variants are grouped to keep the index compact.}
\label{tab:baseline_index}
\scriptsize
\setlength{\tabcolsep}{2.4pt}
\renewcommand{\arraystretch}{0.98}
\begin{tabularx}{\columnwidth}{@{}
  >{\raggedright\arraybackslash}p{0.47\columnwidth}
  >{\raggedright\arraybackslash}p{0.18\columnwidth}
  >{\raggedright\arraybackslash}X@{}}
\toprule
\rowcolor{DriveTeal}
\color{white}\textbf{Baseline(s)} &
\color{white}\textbf{Task} &
\color{white}\textbf{Reported in} \\
\midrule
\rowcolor{DriveMint!38}\multicolumn{3}{@{}l}{\textbf{Representations}} \\
Descriptors (anchor) & re-ID, matched & Tab.~\ref{tab:t2}; Sec.~\ref{sec:results:matched} \\
PatchTST, PatchTST-CI, iTransformer, ArcFace & re-ID, matched & Tabs.~\ref{tab:t2}, \ref{tab:t3full} \\
Masked-TS SSL, JEPA-style SSL, CLIP CAN--video, MOMENT-1 & re-ID, matched & Tabs.~\ref{tab:t2}, \ref{tab:t3full} \\
Qwen3-4B and Llama-3.2-3B text encoders & re-ID & Tab.~\ref{tab:t2}; App.~\ref{app:llm} \\
\rowcolor{DriveMint!38}\multicolumn{3}{@{}l}{\textbf{Shortcut control}} \\
ResidualStyle, DANN, video-only probe & residual / leakage & Secs.~\ref{sec:results:confound}, \ref{sec:results:vehicle}, \ref{sec:results:video}; Tab.~\ref{tab:t3full} \\
\rowcolor{DriveMint!38}\multicolumn{3}{@{}l}{\textbf{Personalization}} \\
FiLM, query injection, re-ID conditioning, vehicle conditioning, Transformer head & prediction (PG) & Secs.~\ref{sec:results:pg}, \ref{sec:results:vehicle}; App.~\ref{app:conditioning} \\
\rowcolor{DriveMint!38}\multicolumn{3}{@{}l}{\textbf{Multimodal scene information}} \\
DINOv2, DINOv3, SigLIP2, V-JEPA~2 & event / prediction & Tab.~\ref{tab:video} \\
Qwen3-VL, Qwen2.5-VL, and Qwen3 text variants & pair / event / maneuver & Apps.~\ref{app:vlm}, \ref{app:llm}, \ref{app:maneuvers} \\
Qwen3-8B + LoRA-SFT & event forecasting & App.~\ref{app:llm} \\
\rowcolor{DriveMint!38}\multicolumn{3}{@{}l}{\textbf{Distributional heads}} \\
MDN, CVAE & prediction (NLL PG) & Sec.~\ref{sec:results:pg} \\
\bottomrule
\end{tabularx}
\end{table}

\section{Baseline Configurations and Metric Implementation}
\label{app:baselines}

All baselines train on the full corpus with AdamW, learning rate $3{\times}10^{-4}$, weight decay $10^{-4}$, and batch size 64--256 on a single RTX 5080 (16\,GB).
Encoders: the joint-patch encoder patches the $600\times17$ window with patch length 10 into 60 tokens (dimension 192, depth 4, attentive statistics pooling, 128-d output); the channel-independent variant processes each channel separately with shared weights (patch 20, dimension 128, depth 3) and pools over channel summaries; iTransformer embeds each channel's full series as one token (dimension 192, depth 4).
Objectives: supervised-contrastive loss with $P{=}16$ drivers $\times$ $K{=}4$ windows per batch and temperature 0.1; ArcFace with scale 30 and margin 0.3.
Channel dropout (15\% of channels, half of the batches) is applied everywhere for missing-channel robustness.
Self-supervised pretraining masks 40\% of patch tokens for 15 epochs before an 8-epoch frozen-trunk probe.
The prediction models are two-layer GRUs (dimension 128) or a three-layer Transformer encoder (Appendix~\ref{app:conditioning}), conditioned through FiLM on a 64-d support encoding; the support encoder is a bidirectional GRU over $k$ support windows.
The multimodal predictor attends from the CAN state to 10 per-frame tokens (2\,fps) or 2--3 temporal clip tokens through single-head cross-attention with a learned gate.
Distributional heads: a 5-component mixture-density network and a conditional VAE with a 16-d latent, KL annealing over 3 epochs, and a decoder log-variance floor of $-3$.

\textbf{Distribution-metric implementation.}
KL divergence uses 30-bin histograms over the pooled 0.5--99.5 percentile range with additive smoothing ($10^{-6}$); MMD uses an RBF kernel with the median-distance bandwidth heuristic; Wasserstein-1 uses the empirical distributions.
All three are computed separately within each scenario bucket (buckets with fewer than 20 samples are skipped) and averaged over buckets and behavior features; the exact code ships with the harness.

\section{Foundation-Model Baselines: LLMs and VLMs}
\label{app:llm}

Beyond the MOMENT-1 time-series foundation model, we evaluate language and vision--language models as zero-shot baselines.
These rows sit \emph{outside} the frozen three-seed protocol of Section~\ref{sec:benchmark:protocol}: they are single-pass, zero-shot evaluations, reported for reference and marked as such wherever they appear.

\textbf{Setup.}
Each window is serialized to compact CSV text over five realized-motion channels (speed, acceleration, curvature$\times 10^3$, yaw rate, THW with 9.9 encoding ``no lead''); 1\,Hz over 60\,s for re-identification, 10\,Hz over the 5\,s history for event forecasting.
The serialization contains no driver, vehicle, or location identifiers.

\textbf{Re-identification (Table~\ref{tab:t2}).}
For the LLM rows, the window embedding is the attention-masked mean of the last hidden layer of a frozen decoder LLM reading the serialized window, evaluated with the identical enrollment protocol and unseen-driver population as every other row.
Qwen3-4B reaches AUROC .596 and Llama-3.2-3B .598 at 5-minute enrollment---at or below the descriptor anchor (.707) and far below trained encoders (.935), matching the MOMENT-1 pattern: generic pretraining, over time series or language, does not encode driver identity without adaptation.

\textbf{Event forecasting (Table~\ref{tab:llmt5}).}
Zero-shot models answer the forecasting question (``will a hard-braking or sharp-steering event begin within 5\,s?'') on a fixed 2{,}897-anchor subsample of the evaluation split (one-third positives); the score is the yes/no first-token log-odds, and the standard CAN GRU is re-scored on the identical anchors.
Every zero-shot model falls well short of the trained 64-unit GRU: the best VLM reaches AUROC .759 against the GRU's .857.
Two details are informative.
The text-only LLM beats the newer VLM given the same CAN serialization (.607 vs.\ .582), and the older Qwen2.5-VL outperforms Qwen3-VL by a wide margin (.759 vs.\ .582)---zero-shot forecasting ability does not track model recency, echoing the per-frame-encoder finding of Section~\ref{sec:results:video}.
Scale does not appear to help either: within the same family, Qwen2.5-VL-7B scores .607 on the identical anchors, well below its 3B sibling (single zero-shot runs, so we read these gaps as calibration-sensitive rather than definitive).
Light adaptation changes the picture: one epoch of LoRA fine-tuning ($r{=}16$, answer-token loss) on the training-fold serializations lifts Qwen3-8B to AUROC .871 / AP .792 on the full 3{,}000-anchor subsample, slightly above the trained GRU on the same protocol (single run, single seed).
Consistent with the representation results of Section~\ref{sec:results:ladder}, generic pretraining carries little zero-shot driving signal but adapts quickly once the task is supervised.
Consistent with that finding, short-horizon anticipation appears to live in vehicle dynamics that a small trained sequence model captures and current foundation models do not extract zero-shot.

\begin{table}[htbp]
\centering
\caption{Foundation models on event forecasting using a fixed anchor subsample with one-third positives (zero-shot rows: the 2{,}897 frames-available anchors; $^{\ddagger}$LoRA row: the full 3{,}000 anchors). These single-pass results are outside the frozen three-seed protocol.}
\label{tab:llmt5}
\footnotesize
\setlength{\tabcolsep}{3.2pt}
\renewcommand{\arraystretch}{1.12}
\begin{tabularx}{\columnwidth}{@{}
  >{\raggedright\arraybackslash}p{0.34\columnwidth}
  >{\raggedright\arraybackslash}X
  >{\centering\arraybackslash}p{0.12\columnwidth}
  >{\centering\arraybackslash}p{0.10\columnwidth}@{}}
\toprule
\rowcolor{DriveTeal}
\color{white}\textbf{Model} &
\color{white}\textbf{Input} &
\color{white}\textbf{AUROC} &
\color{white}\textbf{AP} \\
\midrule
Qwen3-VL-4B (zero-shot) & 3 frames + CAN text & .582 & .413 \\
Qwen3-4B (zero-shot) & CAN text & .607 & .439 \\
Qwen2.5-VL-7B (zero-shot) & 3 frames + CAN text & .607 & .430 \\
Qwen2.5-VL-3B (zero-shot) & 3 frames + CAN text & .759 & .614 \\
\rowcolor{DriveMint!38}
CAN GRU (trained, same anchors) & 5-s CAN history & .857 & .785 \\
\rowcolor{DriveGreen!12}
\textbf{Qwen3-8B + LoRA (1 epoch)$^{\ddagger}$} & \textbf{CAN text} & \textbf{.871} & \textbf{.792} \\
\bottomrule
\end{tabularx}
\end{table}

\begin{table}[htbp]
\centering
\caption{
Comparison of per-frame and temporal video features.
Event forecasting reports CAN-only $\rightarrow$ CAN+video AUROC.
Prediction gain is the relative RMSE improvement from adding video
($k{=}5$; positive values indicate lower RMSE).
}
\label{tab:video}

\setlength{\tabcolsep}{2.8pt}
\renewcommand{\arraystretch}{1.12}
\footnotesize

\begin{tabularx}{\linewidth}{
  >{\raggedright\arraybackslash}X
  >{\centering\arraybackslash}p{0.25\linewidth}
  >{\centering\arraybackslash}p{0.14\linewidth}
  >{\centering\arraybackslash}p{0.21\linewidth}
}
\toprule
\rowcolor{DriveTeal}
\color{white}\textbf{Video encoder} &
\color{white}\textbf{Forecasting AUROC} &
\color{white}\textbf{$\Delta$} &
\color{white}\textbf{Prediction gain} \\
\midrule

\rowcolor{DriveGold!14}
\multicolumn{4}{l}{\textbf{Per-frame appearance features}} \\

DINOv2 '23 &
.835 $\rightarrow$ .787 &
$-.048$ &
$+0.09\%$ \\

\rowcolor{DriveGold!8}
DINOv3 '25 &
.833 $\rightarrow$ .760 &
\textbf{$-.073$} &
$+0.09\%$ \\

SigLIP2 '25 &
.834 $\rightarrow$ .808 &
$-.026$ &
--- \\

\addlinespace[1pt]
\rowcolor{DriveBlue!12}
\multicolumn{4}{l}{\textbf{Temporal motion features}} \\

\rowcolor{DriveGreen!12}
\textbf{V-JEPA~2 '25} &
$.842 \rightarrow \mathbf{.847}$ &
$\mathbf{+.005}$ &
$\mathbf{+1.49 \pm 0.32\%}$ \\

\bottomrule
\end{tabularx}
\end{table}

\section{Conditioning and Recipe Sensitivity}
\label{app:conditioning}

Three checks show which modeling choices move the personalization result and which do not.
First, replacing the GRU predictor with a Transformer encoder leaves the gain positive at every horizon ($+0.3/+0.5/+0.5\%$ at 1/3/5\,s), so the conclusion does not depend on recurrent heads.
Second, moving the same few-shot style vector from FiLM conditioning to query injection turns the gain negative ($-0.7\%$), so the injection mechanism matters more than the backbone.
Third, the multimodal six-way ablation is sensitive to the dropout schedule used to train the shared model: with four independent dropout pathways, only about 39\% of training steps see the full conditioning stack, and the video and driver gains shrink toward zero; with vehicle conditioning always on and the residual pathway enabled after epoch 5, the gains reported in the main text appear.
We release both recipes.
The style-in-query model also showed a late-training loss spike in one run; its number should be read with that caveat.

\section{Component Versions}
\label{app:vintage}

Table~\ref{tab:vintage} lists every pretrained or named component with its release year and role, so the age of the baseline suite can be audited directly.

\begin{table}[htbp]
\centering
\caption{Component versions used in the baseline suite.}
\label{tab:vintage}
\footnotesize
\setlength{\tabcolsep}{3.5pt}
\renewcommand{\arraystretch}{1.12}
\begin{tabularx}{\columnwidth}{@{}
  >{\raggedright\arraybackslash}p{0.40\columnwidth}
  >{\raggedright\arraybackslash}p{0.20\columnwidth}
  >{\raggedright\arraybackslash}X@{}}
\toprule
\rowcolor{DriveTeal}
\color{white}\textbf{Component} &
\color{white}\textbf{Year} &
\color{white}\textbf{Role} \\
\midrule
PatchTST / iTransformer & 2023 / 2024 & sequence encoders \\
SupCon / ArcFace / ProtoNet & 2020 / 2018 / 2017 & metric objectives \\
MOMENT-1 & 2024 & TS foundation model (zero-shot) \\
DINOv2 / DINOv3 / SigLIP2 & 2023 / 2025 / 2025 & per-frame video encoders \\
V-JEPA~2 & 2025 & temporal video encoder \\
Qwen2.5-VL / Qwen3-VL & 2025 / 2025 & scene-attribute annotators \\
FiLM / TransFuser-style injection & 2017 / 2022 & conditioning mechanisms \\
DANN & 2015 & invariance diagnostic \\
MDN / CVAE & classic & distributional heads (by design) \\
\bottomrule
\end{tabularx}
\end{table}

\section{Per-Driver Data Distribution}
\label{app:distribution}

\begin{table}[t]
\centering
\caption{DriveDNA dataset statistics. Driver identifiers are salted-hashed, and the released signals contain no GPS.}
\label{tab:corpus}
\renewcommand{\arraystretch}{1.13}

\begin{tabularx}{\linewidth}{
  >{\raggedright\arraybackslash}X
  >{\raggedleft\arraybackslash}p{0.27\linewidth}
}
\toprule
\rowcolor{DriveTeal}
\color{white}\textbf{Dataset property} &
\color{white}\textbf{Value} \\
\midrule

\rowcolor{DriveMint!38}
\multicolumn{2}{l}{\textbf{Raw corpus}} \\

Drivers & 465 \\
Vehicle models / platform fingerprints & 115 / 216 \\
Drives (raw / decoded) & 4{,}373 / 4{,}121 \\
CAN telemetry / forward video & 1{,}938 / 2{,}027 h \\

\addlinespace[1pt]
\rowcolor{DriveGold!32}
\multicolumn{2}{l}{\textbf{HD data}} \\

HD drivers & 452 \\
HD segments ($\geq$30\,s) & 12{,}440 \\
\rowcolor{DriveGreen!14}
\textbf{Human-controlled hours (total / moving)} &
\textbf{975 / 581 h} \\
Tagged 60\,s windows (drivers) & 62{,}674 (428) \\

\addlinespace[1pt]
\rowcolor{DriveBlue!22}
\multicolumn{2}{l}{\textbf{Annotations and release}} \\

Driving scenarios / behavioral primitives & 6 / 8 \\
Maneuver-event annotations & 276{,}248 \\
Signal / video-embedding rate & 10\,Hz / 2\,fps \\

\bottomrule
\end{tabularx}
\end{table}

\begin{figure*}[htbp]
\centering
\includegraphics[width=\textwidth]{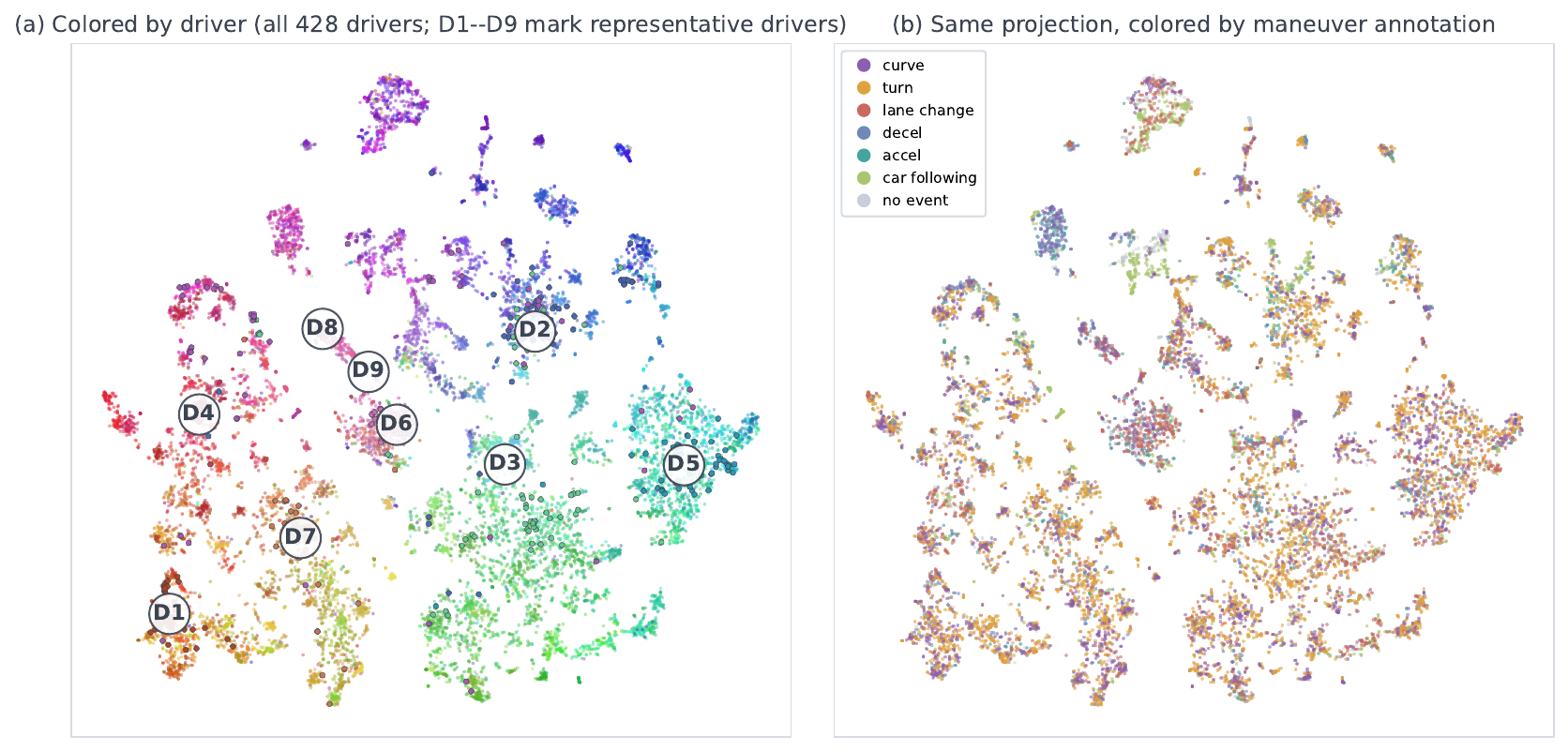}
\caption{t-SNE map of the learned driver-style embedding space (frozen SupCon encoder; at most 60 windows per driver).
\textbf{(a)} All 428 drivers, hue assigned by centroid position; labeled circles mark the nine representative drivers discussed in the text.
\textbf{(b)} The same projection colored by maneuver annotation (rarest class per window).
Clusters align with drivers, not maneuver content.
The encoder is trained on the training fold; the map includes all folds and is intended as a qualitative view.}
\label{fig:embmap}
\end{figure*}

Figure~\ref{fig:dataset_distribution}(b) shows how moving ($v>2$\,m/s) human-driving time is distributed over drivers.
The distribution is heavy-tailed, as expected for a consumer fleet: 449 drivers have moving human-driving data, the median driver contributes about 20 minutes, and the ten largest contributors hold 39\% of all hours.
The benchmark design absorbs this skew in three places.
Few-shot enrollment needs only 1--10 minutes of support data, which the median driver satisfies.
Evaluation folds require at least two drives and ten windows per driver, which removes drivers with too little data for drive-disjoint evaluation.
Training batches sample drivers uniformly ($P \times K$ sampling), so large contributors do not dominate the gradient.

\textbf{A map of the learned driver-style space.}
Figure~\ref{fig:embmap} projects the frozen SupCon window embeddings of all 428 drivers with windows onto two dimensions with t-SNE (at most 60 windows per driver, so that large contributors do not dominate the map).
In panel (a), each driver receives a hue determined by the angular position of its centroid, so neighboring clusters remain distinguishable; windows of the same driver form compact clusters.
Nine representative drivers (D1--D9) are outlined and labeled: eight are selected by clustering driver centroids and taking the best-covered driver of each region (D1: a Hyundai Ioniq~5 driver with 218 windows; D2: a Hyundai Tucson driver observed on five vehicles; D3: a Volkswagen Golf driver; D4: a Kia Sportage driver; D5: a Rivian R1 driver; D6: a Ford F-150 driver; D7: a Hyundai Kona driver; D8: a Ford F-150 Lightning driver), and D9 is the driver observed on the most vehicles (nine, 5{,}378 windows); D9's windows spread over several regions of the map, consistent with the cross-vehicle analysis of Appendix~\ref{app:vehinstance}.
Panel (b) colors the same projection by the maneuver-event annotation of each window (the rarest class among its events, in the priority curve $>$ turn $>$ lane change $>$ decel $>$ accel $>$ car following): maneuver colors mix within driver clusters rather than forming their own, with only mild local structure (30-nearest-neighbor label agreement $1.8\times$ chance, against $130\times$ for driver identity), consistent with the embedding organizing primarily by driver identity rather than by maneuver content.
The projection is qualitative and complements, but does not replace, the quantitative evaluations of Section~\ref{sec:results}.

\section{Datasheet}
\label{app:datasheet}

Following the datasheet framework~\cite{gebru2021datasheets}; the full datasheet and a Croissant metadata record accompany the release.

\textbf{Motivation.}
Created to support rigorous research on personalized driving-style modeling in which vehicle, route, and driving-condition confounding is measurable; no comparable multi-driver, multi-vehicle naturalistic corpus with a driver hierarchy existed.

\textbf{Composition.}
465 drivers, 115 distinct vehicle models, 4{,}121 decoded drives; 975 hours of human-controlled driving as 10\,Hz signal tables with forward video; 62{,}674 tagged windows; over 276{,}000 audited maneuver events; splits and manifests (Section~\ref{sec:dataset}).
The corpus skews toward one region and toward drivers who chose to install an aftermarket driver-assistance system.

\textbf{Collection.}
Recorded between March 2023 and July 2026 by consumer vehicles running openpilot (a small number of drives with unsynchronized device clocks are excluded from this statistic).
Participants provided informed consent for research use and were compensated with gift cards.
No data was collected specifically for this project; the corpus repurposes logs the vehicles already produced.

\textbf{Driver identity.}
Every drive in the corpus carries a driver identifier that uniquely labels the contributing driver; all drives with the same identifier belong to one driver, including drives on different vehicles.
The released data replaces these identifiers with salted hashes.

\textbf{Preprocessing.}
Uniform-tier decode at 10\,Hz; automation-engaged frames removed; driver identifiers salted-hashed; VINs and device identifiers dropped; no GPS in released signals; faces and license plates blurred in the gated video tier.

\textbf{Uses.}
Intended: driving-style representation learning, personalized behavior prediction, confound and robustness research.
Prohibited: attempts to re-identify individuals, surveillance, and insurance or employment decisions about identifiable people.

\textbf{Distribution and maintenance.}
Tiered release (public signals, embeddings, splits, code; data-use agreement for raw video) maintained by the authors, with versioned errata.

\end{document}